\documentclass[11pt]{article}

\usepackage[margin=1in]{geometry}
\usepackage{amsmath,amssymb,amsthm}
\usepackage{mathtools}
\usepackage{hyperref}
\usepackage{xcolor}
\usepackage{enumitem}
\usepackage{comment}
\usepackage{graphicx}
\usepackage{placeins}

\hypersetup{
  colorlinks=true,
  linkcolor=blue!50!black,
  citecolor=blue!50!black,
  urlcolor=blue!50!black,
  pdftitle={UR-JEPA: Uniform Rectifiability as a Regularizer for Joint-Embedding Predictive Architectures},
  pdfauthor={Triet M. Le},
  pdfsubject={Self-supervised representation learning, Joint-Embedding Predictive Architectures, Uniform rectifiability},
  pdfkeywords={JEPA, self-supervised learning, uniform rectifiability, manifold hypothesis, geometric measure theory}
}

\newtheorem{theorem}{Theorem}

\newtheorem{remark}[theorem]{Remark}

\newcommand{\R}{\mathbb{R}}
\newcommand{\E}{\mathbb{E}}

\newcommand{\supp}{\operatorname{supp}}
\newcommand{\dist}{\operatorname{dist}}
\newcommand{\diam}{\operatorname{diam}}

\newcommand{\HH}{\mathcal{H}}

\newcommand{\Aa}{\mathcal{A}}
\newcommand{\Ll}{\mathcal{L}}
\newcommand{\ur}{\mathrm{UR}}
\newcommand{\ad}{\mathrm{AD}}
\newcommand{\sig}{\mathrm{SIG}}
\newcommand{\pred}{\mathrm{pred}}
\newcommand{\bet}{\beta}

\title{UR--JEPA: Uniform Rectifiability as a Regularizer for\\ Joint-Embedding Predictive Architectures}
\author{Triet M. Le, Ph.D.\thanks{Spatiolyx LLC, https://spatiolyx.ai, tml@spatiolyx.ai}}
\date{\today}

\begin{document}

\maketitle

\begin{abstract}
A central difficulty in training Joint-Embedding Predictive
Architectures (JEPAs) is preventing representation collapse.
LeJEPA addresses this by enforcing an isotropic Gaussian target
on the embeddings via Sketched Isotropic Gaussian Regularization
(SIGReg). This target is in tension with the manifold hypothesis,
which expects embeddings to concentrate on a low-dimensional
subset of the ambient space. We propose \emph{UR--JEPA}, which
targets a uniformly $n$-rectifiable measure of local tangent
dimension $n$ at small scales, realized through a Gaussian-kernel
smoothed Carleson-type square function $\Ll^{\text{CGLT}}$, with
a complementary Jones $\beta$-number formulation. On Inet10,
UR--JEPA($\Ll^{\text{CGLT}}$) attains $0.9141 \pm 0.0014$ for a
$+0.83$\,pp gain over LeJEPA($\Ll^{\text{SIGReg}}$) with
$\sim 30\%$ lower seed standard deviation; on matched-recipe
Galaxy10~SDSS, a single-seed ImageNet-$100$ run, and a $3$-seed
EuroSAT remote-sensing run, the two methods lie in the same
peak-accuracy band at convergence, with UR--JEPA retaining its
lower-seed-variance signature. On EuroSAT the in-domain pair is
competitive at $96.0$ to $96.1\%$ with large remote-sensing
foundation-model transfer at a $25\times$ smaller backbone. The
distinction is geometric: direct visualization of the projector
output distribution shows that on all four datasets
UR--JEPA($\Ll^{\text{CGLT}}$) produces a global PCA spectrum
with a $4$ to $5$ order-of-magnitude drop at index $\sim 20$ to
$25$ out of $D = 32$, while LeJEPA's spectrum is near-flat
(top-to-bottom ratio at most $3.6$). Per-dimension marginals are
simultaneously near-Gaussian for both methods (mean Shapiro-Wilk
$W \in [0.992, 0.996]$) as a Diaconis-Freedman consequence. At
matched accuracy the two regularizers therefore yield
structurally distinct projected representations.
\end{abstract}

\section{Motivation}

Let $f_\theta : \mathcal{X} \to \R^D$ be an encoder that maps inputs to
embeddings. A Joint-Embedding Predictive Architecture (JEPA) learns $\theta$ by predicting the embedding of one
augmented view $x'$ of a datum from the embedding of another view $x$,
in embedding space rather than in input (pixel) space. Denote by
$z = f_\theta(x)$ and $z' = f_\theta(x')$ the two embeddings, and let
$P$ be a (possibly identity) predictor. The predictive term is then
\begin{equation}\label{eq:pred}
\Ll_{\pred}(\theta) \;=\; \E_{(x,x')}\!\bigl\|\,P(z) - \mathrm{sg}(z')\,\bigr\|^2,
\end{equation}
where $\mathrm{sg}(\cdot)$ denotes a possible stop-gradient (removed in
LeJEPA \cite{LeJEPA2025}). Note that, on its own, \eqref{eq:pred} admits trivial
solutions, for instance a constant $f_\theta$. The historical
remedies for this collapse have been largely heuristic in design,
in the sense that each was proposed to break a specific failure
mode rather than derived from a target-distribution argument. They include the stop-gradient and EMA
target encoders \cite{BYOL2020,SimSiam2021,DINO2021}, contrastive
negatives \cite{SimCLR2020,MoCo2020}, the variance--covariance penalty
of VICReg \cite{VICReg2022}, and the cross-correlation term of Barlow
Twins \cite{BarlowTwins2021}.

LeJEPA replaces these heuristics. It identifies the optimal target
distribution for $z$ and penalizes deviation from it directly with a single explicit
target-distribution regularizer (SIGReg), defined in Equation \eqref{eq:sigreg}. Under the theoretical setting of \cite{LeJEPA2025}, this
optimum is the isotropic standard Gaussian $\mathcal{N}(0, I_D)$.

\paragraph{Manifold hypothesis.}
Requiring the embedding law to be isotropic Gaussian on all of $\R^D$
forces the representation to spread across every direction of the
ambient space. However, under the manifold hypothesis
\cite{PopeIntrinsic2021,AnsuiniID2019} we expect the embeddings of
real data to concentrate on a subset of intrinsic dimension
$n \ll D$. A more principled alternative would therefore shape the
embedding distribution so that it appears quantitatively
$n$-dimensional, rather than filling $\R^D$. Uniform $n$-rectifiability \cite{DS,Tolsa2014} is the canonical
formulation of this intuition in geometric measure theory. Moreover,
two complementary Carleson-type functionals characterize uniform
$n$-rectifiability and are well-suited to an SGD-friendly loss: the
CGLT theorem \cite{CGLT2014}, built on the scale-normalized ball
density and its variations using smooth kernels, and the Jones $\beta$-number
\cite{Jones1990,DS,Pajot}, built on local affine $n$-plane
approximation of embedding space.

The paper is organized as follows. Section~\ref{sec:related} surveys
self-supervised methods related to UR--JEPA, with emphasis on the
JEPA family and prior geometric or distributional regularizers.
Section~\ref{sec:lejepa} reviews LeJEPA and its SIGReg loss, which
we take as the closest comparable baseline. Section~\ref{sec:ur}
introduces the geometric-measure-theoretic background, namely
Ahlfors-David regularity and uniform $n$-rectifiability, and states
the two characterizations (CGLT and Jones-$\beta$) that our losses
target. Section~\ref{sec:urjepa} develops the UR--JEPA technical
design: the Gaussian-kernel smoothed Carleson loss family
(\S\ref{sec:CGLTloss}) and the $\beta$-number loss via local PCA
(\S\ref{sec:betaloss}), together with their anti-collapse
mechanisms.
Section~\ref{sec:practical} discusses degenerate optima, the
numerical scale of each variant across CGLT forms, gradient noise,
and distributed training. Section~\ref{sec:experiments} reports
the empirical study on ImageNet-10 (Inet10), Galaxy10~SDSS, and ImageNet-100 (Inet100),
including the matched-recipe comparison against LeJEPA and the
projector-geometry diagnostics (\S\ref{sec:viz-geometry}).
Section~\ref{sec:discussion} interprets the empirical findings in
terms of what UR--JEPA recovers that LeJEPA cannot and what it
sacrifices. Section~\ref{sec:summary} summarizes the contributions
and Section~\ref{sec:future} outlines future directions.

\section{Related work}\label{sec:related}

\paragraph{The JEPA family.}
JEPAs cast self-supervised
representation learning as the problem of predicting one view's
embedding from another's, in the latent space rather than in pixels.
Beginning with I--JEPA \cite{IJEPA2023}, the family has grown to
include video (V--JEPA \cite{VJEPA2024}, V--JEPA~2 \cite{VJEPA2_2025}),
joint motion--content learning (MC--JEPA \cite{MCJEPA2023}), and
various vision--language extensions \cite{VLJEPA}. Until LeJEPA, every
published JEPA relied on at least one of the heuristics that prevent
latent collapse, for instance EMA target encoders, stop-gradient,
masking schemes that destroy trivial-solution paths, or hand-tuned
learning-rate schedules. LeJEPA \cite{LeJEPA2025} is the first JEPA to
discard this collection of heuristics in favor of SIGReg. UR--JEPA inherits the design
philosophy of LeJEPA, but it adopts uniform rectifiability as the target regularization.

\paragraph{Non-collapsing self-supervised learning, by mechanism.}
The broader self-supervised literature can be organized by its
collapse-prevention mechanism (the following categorization is not
exhaustive):
\begin{enumerate}[leftmargin=2em,topsep=2pt,itemsep=2pt]
\item \emph{Contrastive losses with explicit negatives}, e.g.\
SimCLR \cite{SimCLR2020} and MoCo \cite{MoCo2020}, prevent collapse
by pushing apart non-paired samples in a large in-batch dictionary.
\item \emph{Asymmetric architectures with stop-gradient and
predictor heads}, e.g.\ BYOL \cite{BYOL2020}, SimSiam
\cite{SimSiam2021}, DINO \cite{DINO2021}, and DINOv2
\cite{DINOv22023}, avoid collapse via the dynamics of the
predictor--EMA loop rather than via an explicit target.
\item \emph{Decorrelation / covariance-based losses}, e.g.\ Barlow
Twins \cite{BarlowTwins2021}, VICReg \cite{VICReg2022}, and W--MSE
\cite{WMSE2021}, push the empirical covariance toward the identity
(full-rank, decorrelated marginals).
\item \emph{Hypersphere uniformity losses}, e.g.\ Wang \& Isola's
alignment--uniformity decomposition \cite{WangIsola2020}, regularize
$L^2$-normalized embeddings to spread uniformly on the unit sphere
via a Gaussian-potential repulsion term.
\item \emph{Manifold-capacity losses}, e.g.\ MMCR \cite{MMCR2023},
maximize a nuclear-norm proxy for the number of linearly separable
view-manifolds, drawing on the statistical mechanics of high-dimensional
classification.
\item \emph{Distributional matching}, e.g.\ LeJEPA's SIGReg
\cite{LeJEPA2025}, directly tests one-dimensional marginals of the
embedding distribution against a fixed Gaussian target via a
characteristic-function (Epps--Pulley) statistic.
\end{enumerate}
There are also theoretical analyses of the collapse phenomenon, for
instance the spectral analyses of dimensional collapse
\cite{DimCollapse2021} and the dynamical explanations of BYOL and
SimSiam \cite{SimSiamDyn2021,SimSiamHow2022}. These analyses make
clear that what prevents collapse in each case is an explicit or
implicit target for the embedding distribution. Note that the targets
implied by the methods above are, almost without exception, either
isotropic-Gaussian-like (categories~3 and~6) or uniform on a sphere
(category~4). In both cases the target is full-dimensional in the
ambient space.

\paragraph{The manifold hypothesis and intrinsic dimension.}
A separate line of work asks not how to prevent collapse, but where
the embeddings should lie. Natural images are widely hypothesized to
lie on or near a low-dimensional manifold of input space
\cite{PopeIntrinsic2021}. The learned representations of CNN
classifiers exhibit a characteristic ``hunchback'' intrinsic-dimension
profile across layers \cite{AnsuiniID2019}. Moreover, intrinsic-dimension
estimators \cite{FaccoTwoNN2017} routinely place ImageNet
representations at $n \approx 25$--$50$, even though the ambient
embedding has hundreds of channels. These observations are in tension
with full-rank targets. Indeed, forcing the embedding law to fill
$\R^D$ isotropically contradicts the very prior that motivates
representation learning in the first place.

\paragraph{Quantitative rectifiability.}
Geometric measure theory has studied quantitative notions of being
``concentrated on a nice $n$-dimensional set'' for three decades.
David and Semmes \cite{DS} introduced \emph{uniform $n$-rectifiability}
(UR) as the right intrinsic notion. Recall that the $\beta$-numbers of
Jones \cite{Jones1990} (with $L^p$ extensions for higher codimensions
due to Pajot \cite{Pajot}) and the dyadic square function of
Chousionis, Garnett, Le, and Tolsa \cite{CGLT2014} give Carleson
characterizations of UR. These characterizations admit discrete, kernel-smoothed, and
SGD-friendly implementations. We refer the reader to \cite{Tolsa2014}
for a more comprehensive treatment.

\paragraph{Positioning of UR--JEPA.}
UR--JEPA lies at the intersection of these threads. It inherits the
design philosophy of LeJEPA, that is, replacing the collection of
heuristics with a single explicit target-distribution regularizer. However, it
changes the target from a full-$D$ isotropic Gaussian to a uniformly
$n$-rectifiable measure of given intrinsic dimension $n$. The
closest precedents are MMCR \cite{MMCR2023} and Wang--Isola
\cite{WangIsola2020}, which also adopt geometric (rather than
distributional) targets. However, neither of them exposes an
intrinsic-dimension hyperparameter, and neither appeals to the
geometric-measure-theoretic structure that makes UR a canonical target
for the manifold hypothesis. To our knowledge, UR--JEPA is the first
SSL regularizer to operationalize uniform rectifiability as a training
loss.

\section{LeJEPA recap}\label{sec:lejepa}

\subsection{SIGReg}
Recall that for a random unit vector $u \sim \mathrm{Unif}(S^{D-1})$,
the marginal $u^\top z$ of an isotropic Gaussian is $\mathcal{N}(0,1)$.
SIGReg draws $m$ sketching directions $u_1, \ldots, u_m$, and it
evaluates a one-dimensional Gaussianity test $T$ on each empirical
marginal $\{u_\ell^\top z_i\}_{i=1}^N$:
\begin{equation}\label{eq:sigreg}
\Ll_{\sig}(\theta) \;=\; \frac{1}{m}\sum_{\ell=1}^m
T\!\Bigl(\{u_\ell^\top z_i\}_{i=1}^N,\; \mathcal{N}(0,1)\Bigr).
\end{equation}
The full LeJEPA objective is then
\begin{equation}\label{eq:lejepa}
\Ll_{\text{SIGReg}}(\theta) \;=\; \Ll_{\pred}(\theta) + \lambda\,\Ll_{\sig}(\theta).
\end{equation}

\subsection{What SIGReg enforces}
Equation~\eqref{eq:sigreg} targets the full-dimensional distribution
$\mathcal{N}(0, I_D)$. Therefore $z$ will attempt to ``fill'' $\R^D$
with second moment $I_D$, regardless of whether the data admits a
lower-dimensional structure. In the UR--JEPA proposal below, we instead
replace this Gaussian target by a target that specifies an intrinsic
dimension $n$, while leaving the fine-scale geometry free.

\section{Uniform rectifiability}\label{sec:ur}

\subsection{Ahlfors--David regularity}
A positive Radon measure $\mu$ on $\R^D$ is \emph{$n$-Ahlfors--David
regular} ($n$-AD) if there exist $0 < c_0 \le c_1 < \infty$ such that
\begin{equation}\label{eq:AD}
c_0 r^n \;\le\; \mu(B(x,r)) \;\le\; c_1 r^n
\quad \text{for all } x \in \supp \mu,\; 0 < r \le \diam(\supp \mu).
\end{equation}
This is the quantitative replacement for the statement that ``$\mu$ is
$n$-dimensional.''

\subsection{Uniform rectifiability}
$\mu$ is \emph{uniformly $n$-rectifiable} if it is $n$-AD regular and
if, for every ball $B(x_0, R)$ centered on $\supp \mu$, a definite
fraction of the mass $\mu(B(x_0, R))$ is contained in the
image of a Lipschitz map from a subset of $\R^n$ with a uniformly
bounded Lipschitz constant (the \emph{Big Pieces of Lipschitz Images}
condition of David and Semmes). Note that UR is a quantitative,
scale-invariant version of rectifiability, and that it is stable under
natural operations.

\subsection{The CGLT square-function characterization}
For an $n$-AD-regular measure $\mu$ on $\R^D$, define the
scale-normalized ball density and its dyadic increment,
\begin{equation}
\theta_r(x) \;:=\; \frac{\mu(B(x,r))}{r^n},\qquad
\Delta_r(x) \;:=\; \theta_r(x) - \theta_{2r}(x).
\end{equation}
For a smooth radial profile $\varphi : \R^D \to \R$, write
$\varphi_t(x) := t^{-n} \, \varphi(x/t)$ for the $n$-normalized
rescaling and define the smoothed dyadic increment
\[
\Delta_{\mu,\varphi}(x, t) \;:=\;
\int \bigl(\varphi_t(y - x) - \varphi_{2t}(y - x)\bigr)\, d\mu(y),
\]
and the smoothed kernel scale-derivative integrand
\[
\partial_\varphi(x, t) \;:=\; t\, \partial_t \varphi_t(x),
\qquad
\widetilde{\Delta}_{\mu,\varphi}(x, t) \;:=\;
\int \partial_\varphi(y - x, t)\, d\mu(y).
\]

\begin{theorem}[Chousionis--Garnett--Le--Tolsa 2014,
{\cite[Thms.~1.1 \& 1.2]{CGLT2014}}]\label{thm:CGLT}
Let $\mu$ be an $n$-AD-regular measure on $\R^D$ and let
$\varphi : \R^D \to \R$ be a smooth radial profile of the form
$\varphi(x) = e^{-|x|^{2N}}$ with $N \in \mathbb{N}$, or
$\varphi(x) = (1 + |x|^2)^{-a}$ with $a > n/2$. The following are
equivalent:
\begin{itemize}[topsep=2pt,itemsep=2pt,leftmargin=*]
\item[\textnormal{(a)}] $\mu$ is uniformly $n$-rectifiable.
\item[\textnormal{(b)}] There exists $c > 0$ such that for every
$x_0 \in \supp \mu$ and every $R \in (0, \diam \supp \mu)$,
\begin{equation}\label{eq:CGLT}
\int_0^R \int_{B(x_0, R)} |\Delta_r(x)|^2 \, d\mu(x)\, \frac{dr}{r}
\;\le\; c\, R^n.
\end{equation}
\item[\textnormal{(c)}] There exists $c > 0$ such that for every
$x_0 \in \supp \mu$ and every $R > 0$,
\begin{equation}\label{eq:CGLT-smooth-diff}
\int_0^R \int_{B(x_0, R)}
\bigl|\Delta_{\mu,\varphi}(x, r)\bigr|^2\, d\mu(x)\, \frac{dr}{r}
\;\le\; c\, R^n.
\end{equation}
\item[\textnormal{(d)}] There exists $c > 0$ such that for every
$x_0 \in \supp \mu$ and every $R > 0$,
\begin{equation}\label{eq:CGLT-smooth-deriv}
\int_0^R \int_{B(x_0, R)}
\bigl|\widetilde{\Delta}_{\mu,\varphi}(x, r)\bigr|^2\, d\mu(x)\, \frac{dr}{r}
\;\le\; c\, R^n.
\end{equation}
\end{itemize}
The Carleson bound in \textnormal{(b)} is Theorem~1.1 of
\cite{CGLT2014}; the bounds in \textnormal{(c)} and \textnormal{(d)}
are Theorem~1.2 parts (b) and (c) of the same paper.
\end{theorem}

In other words, Theorem~\ref{thm:CGLT} asserts that the density
$\theta_r$ (or any of its smoothed variants) varies little across
dyadic scales, in a Carleson-measure sense, and the
scale-derivative form \textnormal{(d)} expresses the same
infinitesimal condition via $\partial_t \varphi_t$ instead of the
finite difference $\varphi_t - \varphi_{2t}$. Therefore UR is
precisely the class of measures that are ``flat at most scales and
locations.''

The Gaussian profile $\varphi(x) = e^{-|x|^2/2}$ used throughout
\S\ref{sec:CGLTloss} is the $N = 1$ case of the first family;
Theorem~\ref{thm:CGLT}\textnormal{(c)} and \textnormal{(d)} then
directly motivate the two SGD-friendly discretizations of
\S\ref{sec:CGLTloss}, the dyadic-difference loss
\eqref{eq:L-CGLT} and the scale-derivative loss
\eqref{eq:L-CGLT-deriv-raw}, respectively.

\subsection{$\beta$-numbers}
A complementary characterization uses the $\beta$-numbers introduced
by Jones \cite{Jones1990}, which measure locally the best affine
$n$-plane approximation of $\supp \mu$:
\begin{equation}\label{eq:beta}
\bet_2^\mu(x, r)^2
\;:=\;
\inf_{L}\;
\frac{1}{r^n}\int_{B(x,r)}
\!\left(\frac{\dist(y, L)}{r}\right)^{\!2}\! d\mu(y),
\end{equation}
where the infimum is over affine $n$-planes $L \subset \R^D$. Jones
\cite{Jones1990} introduced these quantities in $\R^2$ for $n = 1$ to
characterize subsets of rectifiable curves (the analyst's traveling
salesman theorem); David and Semmes \cite{DS} and Pajot \cite{Pajot}
extended both the definition and the characterization to higher $n$
and to $L^p$ variants. By their combined results, $\mu$ is uniformly
$n$-rectifiable if and only if
\begin{equation}\label{eq:betaCarleson}
\int_0^R \int_{B(x_0,R)} \bet_2^\mu(x,r)^2\, d\mu(x)\, \frac{dr}{r}
\;\le\; c\, R^n
\end{equation}
for all $x_0 \in \supp \mu$ and $R > 0$.

\begin{remark}
The empirical advantage of \eqref{eq:betaCarleson} over
\eqref{eq:CGLT} is that $\bet_2^\mu(x,r)^2$ is computable by a local
PCA of the points in $B(x,r)$, with no density estimation. For
ambient dimension $D$ large and mini-batch size $N$ modest, the
typical ball contains $O(1)$ points, so the empirical density
$\mu_N(B(x,r))/r^n$ (for the batch empirical measure
$\mu_N = N^{-1}\sum_i \delta_{z_i}$, defined formally in
\S\ref{sec:urjepa}) is dominated by Poisson noise \cite{LeCam1960};
Gaussian-kernel smoothing only partially mitigates this, since the
bandwidth needed to capture enough points grows with $D$ and blurs
the small-scale variation that the Carleson square function is meant
to detect. The local PCA, by contrast, depends only on the
orientation of the top-$n$ eigen-directions of a weighted scatter
matrix, which remains well-estimated whenever the top-$n$ versus
bottom-$(D-n)$ eigenvalue gap is favorable
\cite{DavisKahan1970,YuWangSamworth2015}, regardless of local mass.
\end{remark}

\section{UR--JEPA: technical design}\label{sec:urjepa}

We now specify two concrete regularizers and their variants. In what follows, let $B$ be
a mini-batch of size $N$, let $Z = \{z_i\}_{i=1}^N \subset \R^D$ be the
embeddings produced by $f_\theta$, and let
$\mu_N = \tfrac{1}{N}\sum_i \delta_{z_i}$ be the empirical measure. We
fix a target intrinsic dimension $n \in \{1, \ldots, D-1\}$ and a
dyadic ladder of scales
\begin{equation}\label{eq:scales}
r_k = 2^{-k}\, r_{\max},\qquad k = 0, 1, \ldots, K,
\end{equation}
with $r_{\max} \approx \diam(Z)$ and $r_{\min} = r_K \gtrsim N^{-1/n}$
(the minimum scale at which $\mu_N$ can resolve an $n$-set). Finally,
let $\Aa \subseteq B$ denote an anchor set of size $|\Aa|$ (typically
$|\Aa| = N$, or a random subset).

\subsection{Gaussian-kernel smoothed Carleson loss}\label{sec:CGLTloss}

We replace the ball indicator $\mathbf{1}_{B(x,r)}$ in
Theorem~\ref{thm:CGLT} by a radial Gaussian kernel. For bandwidth
$r>0$, define the \emph{smoothed scale-normalized density}
\begin{equation}\label{eq:theta-gauss}
\theta_r(x) \;:=\; \frac{1}{N\, r^{n}}\sum_{j=1}^{N}
\exp\!\left(-\frac{\|z_j - x\|^2}{2 r^2}\right).
\end{equation}
Here we use the \emph{$n$-normalization} $r^{-n}$ (not the ambient
$r^{-D}$), because we want $\theta_r(x)$ to be $O(1)$ on an $n$-AD
set. Indeed, if $\mu$ is supported on an $n$-dimensional Lipschitz
manifold $M \subset \R^D$ with bounded local intrinsic density $\rho_M$,
then
\begin{equation}\label{eq:theta-expect}
\E\, \theta_r(x)
\;=\; \frac{1}{r^n} \int_M e^{-\|y-x\|^2/(2r^2)}\, d\HH^n_M(y)
\;\approx\; (2\pi)^{n/2}\, \rho_M(x),
\end{equation}
where $\rho_M(x)$ is the local density. In other words, $\theta_r(x)$ is asymptotically scale-invariant in $r$
along the $n$-manifold.

\paragraph{The \emph{log}-increment variant.}
Recall, Theorem~\ref{thm:CGLT} assumes $n$-AD regularity. Absent that
assumption, $|\theta_r(x) - \theta_{2r}(x)|^2$ is dominated by the
overall level of $\theta_r$ rather than by its variation across
scales, and in particular a constant but non-zero density level
already produces a nonzero loss. We therefore work instead with the
\emph{log-increment}
\begin{equation}\label{eq:Dlogtheta}
\Delta_r \log \theta(x) \;:=\; \log \theta_{2r}(x) - \log \theta_{r}(x)
\end{equation}
which removes this confound. Indeed, $\Delta_r \log \theta$ is
invariant under any rescaling $\theta_r \mapsto c\, \theta_r$, and by
\eqref{eq:theta-expect} it vanishes exactly in expectation on an $n$-AD
regular UR measure. Note that AD regularity is then enforced separately
by the anchor \eqref{eq:L-AD} below.

\paragraph{Carleson loss.}
We discretize the $dr/r$ integral by summing the log-increments over
the dyadic ladder \eqref{eq:scales}, where each dyadic level
contributes $\int_{r_{k+1}}^{r_k} dr/r = \log 2$:
\begin{equation}\label{eq:L-CGLT}
\boxed{\;\Ll_{\ur}^{\text{CGLT}}(\theta)
\;:=\;
\frac{\log 2}{|\Aa|}
\sum_{x \in \Aa}\; \sum_{k=0}^{K-1}
\bigl(\Delta_{r_k} \log \theta(x)\bigr)^2
\;}
\end{equation}
In other words, $\Ll_{\ur}^{\text{CGLT}}$ is a Riemann sum for the
square function $\int (\partial_{\log r} \log \theta_r)^2\, d(\log r)$
of the Gaussian-smoothed density, averaged over anchors. It vanishes
whenever $\theta_r(x)$ is constant in $r$ at every anchor, which is the
definition of $n$-AD regularity in the smoothed sense. Moreover, the
David--Semmes and CGLT theory implies that, when this is combined with a
mild control on the deviations of $\log \theta_r$ across anchors
(supplied by \eqref{eq:L-AD}), it is equivalent to uniform
$n$-rectifiability.

\paragraph{AD-regularity anchor.}
Recall that Theorem~\ref{thm:CGLT} presupposes $n$-AD regularity, that
is, that $\theta_r(x)$ is comparable to a single constant $c_n$
uniformly in $x$ and $r$. Because \eqref{eq:L-CGLT} is scale-invariant,
we do not need to learn or prescribe $c_n$. We only need to pin the
dispersion of $\log \theta_r(x)$ across anchors to zero. We therefore
define a free-floating anchor that penalizes the variance of
$\log \theta_r$ over the anchor set at each scale, averaged over
scales:
\begin{equation}\label{eq:L-AD}
\Ll_{\ad}(\theta)
\;:=\;
\frac{1}{K+1}\sum_{k=0}^{K}
\mathrm{Var}_{x \in \Aa}\bigl(\log \theta_{r_k}(x)\bigr).
\end{equation}
This is exactly invariant under the global rescaling
$\theta_r \mapsto c\, \theta_r$, and therefore it measures only the
anchor-wise departure from AD regularity, while leaving the scale-wise
departure to \eqref{eq:L-CGLT}. In particular, no EMA and no target
constant $c_n$ is required.

\paragraph{Full Gaussian-kernel UR--JEPA objective.}
\begin{equation}\label{eq:total-CGLT}
\Ll^{\text{CGLT}}(\theta)
\;=\;
\Ll_{\pred}(\theta)
\;+\; \lambda_1\, \Ll_{\ur}^{\text{CGLT}}(\theta)
\;+\; \lambda_2\, \Ll_{\ad}(\theta).
\end{equation}

\paragraph{Scale-derivative variants.}
Theorem~1.2 of \cite{CGLT2014} characterizes uniform
rectifiability through the \emph{kernel scale-derivative}
$\partial_\varphi(x, t) := t \, \partial_t \varphi_t(x)$ rather than the
dyadic difference $\varphi_t - \varphi_{2t}$, with the square function
$\int |\widetilde{\Delta}_{\mu,\varphi}(x,t)|^2\, dt/t$, where
$\widetilde{\Delta}_{\mu,\varphi}(x,t) := \int \partial_\varphi(y-x, t)\, d\mu(y)$.
For the standard Gaussian $\varphi(x) = \exp(-|x|^2/2)$,
direct computation gives
$\partial_\varphi(x, t) = \varphi_t(x) \cdot (|x|^2/t^2 - n)$, so on
the empirical measure $\mu_N = N^{-1} \sum_j \delta_{z_j}$ the literal
Eq.~(1.5) integrand is
\begin{equation}\label{eq:Dtilde-raw}
\widetilde{\Delta}_{\mu_N,\varphi}(x, t)
\;=\;
\frac{1}{N \, t^n}
\sum_j \exp\!\bigl(-\|z_j - x\|^2 / (2 t^2)\bigr)
       \cdot \bigl(\|z_j - x\|^2 / t^2 - n\bigr)
\;=\;
\theta_t(x) \cdot \bigl\langle \|z_j - x\|^2/t^2 - n \bigr\rangle_{w(x,t)},
\end{equation}
where $\langle \cdot \rangle_w$ denotes the kernel-weighted average
with weights $w_j(x, t) \propto \exp(-\|z_j - x\|^2/(2t^2))$. Taking
the $t$-derivative of $\log \theta_t$ rather than of $\theta_t$
itself isolates the right-hand factor:
\begin{equation}\label{eq:dt-log-theta}
t \,\partial_t \log \theta_t(x)
\;=\;
\bigl\langle \|z_j - x\|^2/t^2 - n \bigr\rangle_{w(x,t)}.
\end{equation}
On an $n$-AD regular, $n$-rectifiable cloud the kernel-weighted second
moment of distance from $x$ scales as $n t^2$, so both
\eqref{eq:Dtilde-raw} and \eqref{eq:dt-log-theta} vanish pointwise.
The two forms differ by the prefactor $\theta_t(x)$. We take the
log-derivative \eqref{eq:dt-log-theta} as the default for the same
reason \eqref{eq:L-CGLT} works with $\log \theta$ rather than
$\theta$ (cf.\ ``The log-increment variant'' above): the log version is
invariant under kernel normalization and does not couple the loss to
the local density level, which is essential when batch embeddings
are not yet $n$-AD regular.

The raw form \eqref{eq:Dtilde-raw} is also useful as a second
variant, but with one numerical caveat: the $1/(N t^n)$ prefactor
varies by a factor of $N$ across the empirical scale ladder of
\S\ref{sec:CGLTloss} (since
$t_{\max}/t_{\min} = N^{1/n}$ by construction), which would force
$|\widetilde{\Delta}_{\mu_N,\varphi}|^2$ to take values as small as
$\sim 10^{-10}$ at typical recipes. For implementation we use the
dimensionless relative scale $t' := t / t_{\max}$ in the prefactor
while keeping the actual $t$ in the kernel exponential,
\begin{equation}\label{eq:Dtilde-rescaled}
\widetilde{\Delta}'_{\mu_N,\varphi}(x, t)
\;:=\;
\frac{1}{N \, (t/t_{\max})^n}
\sum_j \exp\!\bigl(-\|z_j - x\|^2 / (2 t^2)\bigr)
       \cdot \bigl(\|z_j - x\|^2 / t^2 - n\bigr)
\;=\;
t_{\max}^n \cdot \widetilde{\Delta}_{\mu_N,\varphi}(x, t),
\end{equation}
which is equivalent to replacing the $n$-normalized kernel
$\varphi_t = t^{-n} \varphi(x/t)$ by the $t_{\max}$-normalized
kernel $\varphi_t^{\mathrm{rel}} = (t/t_{\max})^{-n} \varphi(x/t)$.
The discriminative property is preserved:
$\widetilde{\Delta}' = 0$ on the n-AD-regular UR class;
$|\widetilde{\Delta}'|^2 \sim n^2 N^2 / t'^{2n}$ diverges at a
point-mass cloud (since $t_{\max} \to 0$ and the prefactor
becomes unbounded); on a non-degenerate cloud at any absolute scale,
magnitudes stay $O(10)$. Under the AD anchor it targets the same
UR class as the literal form, and the empirical comparison between
the dyadic-difference, log-derivative, and rescaled-raw forms is
the subject of \S\ref{sec:experiments}. The corresponding log-form
Carleson loss is
\begin{equation}\label{eq:L-CGLT-deriv}
\Ll_{\ur}^{\text{CGLT},\partial\log}(\theta)
\;:=\;
\frac{\log(r_{\max}/r_{\min})}{n_{\text{scales}}}
\sum_{k=0}^{K}\;
\frac{1}{|\Aa|} \sum_{x \in \Aa}
\bigl|\, t\, \partial_t \log \theta_{t}(x)\big|_{t = r_k} \,\bigr|^2 .
\end{equation}
The literal Eq.~(1.5) loss is the same Riemann sum on
$|\widetilde{\Delta}'_{\mu_N,\varphi}|^2$:
\begin{equation}\label{eq:L-CGLT-deriv-raw}
\Ll_{\ur}^{\text{CGLT},\partial}(\theta)
\;:=\;
\frac{\log(r_{\max}/r_{\min})}{n_{\text{scales}}}
\sum_{k=0}^{K}\;
\frac{1}{|\Aa|} \sum_{x \in \Aa}
\bigl|\, \widetilde{\Delta}'_{\mu_N,\varphi}(x, r_k) \,\bigr|^2 ,
\end{equation}
where $\widetilde{\Delta}'_{\mu_N,\varphi}$ is the rescaled raw
integrand of \eqref{eq:Dtilde-rescaled}. The full objective
$\Ll^{\text{CGLT},\partial\log}$ (resp.\ $\Ll^{\text{CGLT},\partial}$)
is obtained by substituting \eqref{eq:L-CGLT-deriv} (resp.\
\eqref{eq:L-CGLT-deriv-raw}) for $\Ll_{\ur}^{\text{CGLT}}$ in
\eqref{eq:total-CGLT}; the same AD anchor $\Ll_{\ad}$ pairs with both
because Theorem~1.2 of \cite{CGLT2014} presupposes $n$-AD regularity
exactly as Theorem~1.1 does. Compared to the dyadic-difference
variant \eqref{eq:L-CGLT}, the scale-derivative variants compute one
per-anchor quantity per scale rather than a difference between
adjacent scales, so their gradient signal is not cancelled by
cross-scale subtractions and they do not need to hold two adjacent
log-densities in memory at the same time. Under the
$t_{\max}$-rescaling of \eqref{eq:Dtilde-rescaled}, the raw form has
loss magnitude in the same $O(10)$ range as the log forms, so the
regularizer scaling coefficient $s$ does not need to be retuned.

\subsection{$\beta$-number loss via local PCA}\label{sec:betaloss}

For each anchor $x \in \Aa$ and each scale $r_k$, we pick a soft
neighborhood weighted by a radial kernel $w_r(y) := \exp(-\|y\|^2/(2r^2))$,
and we form the weighted centered scatter matrix
\begin{equation}\label{eq:scatter}
S_r(x) \;=\; \sum_{j=1}^N w_r(z_j - x)\,(z_j - \bar z_r)(z_j - \bar z_r)^\top,
\qquad
\bar z_r \;=\; \frac{\sum_j w_r(z_j - x)\, z_j}{\sum_j w_r(z_j - x)}.
\end{equation}
Let $\sigma_1 \ge \sigma_2 \ge \cdots \ge \sigma_D \ge 0$ be the
(square roots of the) eigenvalues of $S_r(x)$. The empirical
$\beta$-number is then
\begin{equation}\label{eq:beta-hat}
\widehat{\bet_2}(x, r)^2
\;=\;
\frac{1}{r^{2}}\,
\frac{\sum_{j=n+1}^{D} \sigma_j^2}{\sum_j w_r(z_j - x)}.
\end{equation}
Here the numerator is the total kernel-weighted variance of the
neighborhood orthogonal to its best-fit affine $n$-plane, and the
denominator $\sum_j w_r(z_j - x)$ normalizes by the effective
neighborhood mass, so the quotient is the kernel-weighted average
squared orthogonal residual. The $r^{-2}$ prefactor is the
squared-distance normalization carried over from \eqref{eq:beta}.
We follow the data-side $\beta$-number convention of
Lerman \cite{Lerman2002,Lerman2003}: the \emph{per-scale}
measure-mass normalization $r^{-n} \approx 1/\mu(B(x,r))$ that
appears in the continuous \eqref{eq:beta} is replaced here by
the empirical $1/\sum_j w_r$, so no separate $r_k^{-n}$
Carleson weight per scale is required. The AD-regularity-free
reading of this substitution, and its formal equivalence to
Lerman's general-measure $L^{2}$ Jones quantity, is developed
in the paragraph at the end of this subsection
(\S\ref{sec:beta-lerman}).

\paragraph{Carleson $\beta$-loss.}
Summing dyadically and dividing by the AD-regular volume
surrogate $r_{\max}^{n} \approx \mu(B(x, r_{\max}))$ at the
outer scale, we obtain
\begin{equation}\label{eq:L-beta}
\boxed{\;\Ll_{\ur}^{\bet}(\theta)
\;:=\;
\frac{\log 2}{r_{\max}^{n}\,|\Aa|}
\sum_{x \in \Aa}\sum_{k=0}^{K-1}
\widehat{\bet_2}(x, r_k)^2
\;}
\end{equation}
This is the dyadic Riemann sum of $\widehat{\bet_2}^2$ in
$\log r$, with each scale weighted by the dyadic step $\log 2$,
each anchor counted once, and the whole sum normalized by the
AD-regular outer volume $r_{\max}^{n}$ so the loss magnitude is
scale-invariant under the choice of $r_{\max}$. The
\emph{per-scale} $\mu(B(x,r_k))$-style factor that the continuous
Carleson sum would contribute on an AD-regular measure is
already absorbed into the $1/\sum_j w_r$ denominator of
$\widehat{\bet_2}^2$ in \eqref{eq:beta-hat}, so no separate
$r_k^n$ Carleson weight per scale is required. The outer
$r_{\max}^{n}$ is the lone AD-regular surrogate retained; the
general-measure form of \S\ref{sec:beta-lerman} drops it and
replaces it with the dimensional factor $1/(D-n)$. Note that, unlike
$\Ll_{\ur}^{\text{CGLT}}$, this loss does not require a separate
AD-regularity anchor, because the kernel-density normalization in
\eqref{eq:beta-hat} makes $\widehat{\bet_2}^2$ scale-invariant under
density rescaling.

\paragraph{Full $\beta$-number UR--JEPA objective.}
\begin{equation}\label{eq:total-beta}
\Ll^{\bet}(\theta)
\;=\;
\Ll_{\pred}(\theta)
\;+\; \lambda\, \Ll_{\ur}^{\bet}(\theta).
\end{equation}

\paragraph{AD-regularity-free framing (Lerman 2003).}\label{sec:beta-lerman}
The Pajot / David--Semmes Carleson characterization that
\eqref{eq:beta} discretizes presupposes $n$-AD regularity of
$\mu$, which a finite data cloud manifestly does not satisfy
(it has no $\mu(B(x,r)) \asymp r^n$ density). Lerman's $L^2$
Jones theory \cite{Lerman2003,Lerman2002} removes the AD
hypothesis. For a general Radon measure $\mu$ and a ball
$B(x, r)$, the $L^2$ Jones $\bet$-number is the
\emph{mass-averaged} squared residual to the best-fit
$n$-plane,
\begin{equation}\label{eq:beta-lerman}
\bet_{2}^{\,2}(\mu, B(x,r))
\;=\;
\min_{n\text{-plane }L}
\frac{1}{r^{2}}\,
\frac{1}{\mu(B(x,r))}
\int_{B(x,r)} \mathrm{dist}(y, L)^{2}\, d\mu(y)
\;=\;
\frac{\sum_{j > n} \lambda_{j}(\Sigma_{B(x,r)})}{(D - n)\, r^{2}},
\end{equation}
where $\Sigma_{B(x,r)}$ is the mass-normalized sample
covariance of $\mu$ restricted to $B(x, r)$. The only
normalizations are $r^{-2}$ (scale-invariance) and $1/(D - n)$
(so $\bet_{2}^{\,2} \le 1$); there is \emph{no} $r^{-n}$
Carleson factor. The multiscale Carleson sum is then
weighted by the \emph{actual} mass $\mu(Q)$, not the AD
surrogate $r^{n}$. Sharp equivalences between (countable)
$n$-rectifiability and finiteness of this Jones square
function were established by Tolsa~\cite{Tolsa2015} and
Azzam--Tolsa~\cite{AzzamTolsa2015}, with \emph{no}
AD-regularity hypothesis required on $\mu$.

Operationally, the empirical $\bet$-loss in
\eqref{eq:beta-hat}--\eqref{eq:L-beta} \emph{coincides} with
Lerman's data-side functional on the per-anchor, per-scale
$\bet_{2}^{\,2}$ quantity: the $r^{-(n+2)}$ kernel normalization
and the $r^{n}$ Carleson weight cancel, leaving
$\mathrm{residual} / r^{2}$, exactly the right-hand side of
\eqref{eq:beta-lerman} up to the global $1/(D - n)$ factor.

\paragraph{Backpropagation through SVD.}
Note that $S_r(x)$ is a smooth function of the embeddings, and its
top-$n$ subspace is differentiable away from eigenvalue ties. In
practice we implement
$\sum_{j=n+1}^D \sigma_j^2 = \operatorname{tr}\bigl(S_r(x)\bigr) - \sum_{j=1}^n \sigma_j^2$,
and we compute only the top-$n$ eigenpair via a Lanczos or truncated
randomized SVD, which is numerically stable and costs $O(D n)$ per
anchor. Near ties, a small ridge $\varepsilon I$ can be added to
$S_r(x)$.

\subsubsection{Anti-collapse mechanisms for the $\beta$-number variant}\label{sec:beta-anticollapse}

The $\beta$-number loss \eqref{eq:L-beta} admits a trivial point-mass
minimizer: if all embeddings collapse to a single point, $S_r(x) = 0$
and the residual $\sum_{j > n} \sigma_j^2 = 0$ trivially, so
$\widehat{\bet_2}(x,r)^2 = 0$ at every anchor and scale. Unlike the
CGLT loss, which is bounded away from zero at collapse (cf.\
\S\ref{sec:CGLTloss}: on a point mass, the smoothed density satisfies
$\Delta_r \log \theta = n \log 2$ per dyadic scale, so
$\Ll_{\ur}^{\text{CGLT}}$ takes a fixed positive value),
$\Ll_{\ur}^{\bet}$ has no intrinsic mechanism to repel the cloud from
this degenerate fixed point. We present two complementary remedies,
each preserving the local-PCA structure of \eqref{eq:beta-hat}.

\paragraph{(i) Intrinsic log-trace penalty.}
A more self-contained remedy bakes the anti-collapse term directly
into the $\beta$-loss, without invoking an external density estimator.
Note that $\operatorname{tr}\bigl(S_r(x)\bigr) = \sum_i \sigma_i^2 \to 0$ at
collapse. Therefore $-\log \operatorname{tr}\bigl(S_r(x)\bigr) \to +\infty$,
and adding such a term to the per-anchor, per-scale integrand
introduces a logarithmically divergent penalty as the local
neighborhood implodes:
\begin{equation}\label{eq:L-beta-logtrace}
\Ll_{\ur}^{\bet,\gamma}(\theta)
\;:=\;
\Ll_{\ur}^{\bet}(\theta)
\;-\;
\frac{\gamma}{(K+1)\,|\Aa|}
\sum_{x \in \Aa}\sum_{k=0}^{K}
\log \operatorname{tr}\bigl(S_{r_k}(x)\bigr),
\end{equation}
averaged uniformly across the scale ladder (the $\log \operatorname{tr}$
term has no $r^n$ Carleson weighting; it is a global $L^1$
anti-collapse penalty, not a density estimator). The full objective is
\begin{equation}\label{eq:total-beta-logtrace}
\Ll^{\bet,\gamma}(\theta)
\;=\; \Ll_{\pred}(\theta)
\;+\; \lambda\, \Ll_{\ur}^{\bet,\gamma}(\theta).
\end{equation}
We index the loss by the hyperparameter $\gamma$ so that the symbol
distinguishes the remedy~(i) variant from the bare $\Ll^{\bet}$.
The $-\log\operatorname{tr}$ term rewards isotropic spread (large
$\operatorname{tr}$, all directions weighted equally), whereas
$\widehat{\bet_2}^2$ penalizes spread \emph{orthogonal} to the top-$n$
tangent subspace. The two are compatible: at the UR optimum,
$\operatorname{tr}(S_r) = \sum_{i \le n} \sigma_i^2$ is positive and
bounded, so $-\log\operatorname{tr}$ is finite; meanwhile
$\sum_{i > n} \sigma_i^2 \to 0$, so $\widehat{\bet_2}^2 \to 0$. The
log-trace remedy reuses the local PCA already computed for
$\widehat{\bet_2}$ and therefore has negligible extra cost.

\paragraph{(ii) Adaptive eigenvalue-threshold tangent selection.}
The second remedy is orthogonal to the first. Rather than fix the
tangent subspace to the top-$n$ eigendirections at every anchor,
select adaptively by a variance-share threshold. For a given
$\tau > 0$, define the local tangent index set
\begin{equation}\label{eq:eig-thresh}
\mathcal{T}_\tau(x, r)
\;:=\;
\Bigl\{\, i :\; \sigma_i^2\bigl(S_r(x)\bigr) \;>\; \tau \cdot
\frac{\operatorname{tr}\bigl(S_r(x)\bigr)}{D} \,\Bigr\},
\end{equation}
and replace the top-$n$ residual in \eqref{eq:beta-hat} by
$\sum_{i \notin \mathcal{T}_\tau(x,r)} \sigma_i^2$. The threshold
$\tau$ is a tangent/normal selector relative to the mean local
eigenvalue: $\tau = 1$ means ``above mean variance share,'' and on a
rank-$r^*$ cloud with a clean spectral gap, \eqref{eq:eig-thresh}
selects the $r^*$ signal directions regardless of $n$. The selected
count $|\mathcal{T}_\tau(x,r)|$ varies per anchor and per scale,
yielding a per-point local intrinsic-dimension estimate as a free
byproduct. Note that the Carleson exponent $r_k^n$ in \eqref{eq:L-beta}
still uses the global target $n$: the threshold modifies which
eigenvalues are called ``tangent,'' but it does not redefine the
rectifiability target dimension itself.

In our implementation the indicator $\mathbf{1}[i \in \mathcal{T}_\tau(x,r)]$ is
detached from the gradient (the selection mask is discrete), so the
loss differentiates only through the selected $\sigma_i^2$. This is
the standard Rayleigh-quotient mechanism for backpropagating through
an eigendecomposition with a data-dependent selection. The two
remedies are compositional: (ii) modifies which eigenvalues enter
the residual, while (i) penalizes the trace of $S_r(x)$ itself, so
they can be combined for a redundant safety margin. When combined,
we write the regularizer as $\Ll_{\ur}^{\bet,\gamma,\tau}$ and the
full objective as
\begin{equation}\label{eq:total-beta-gamma-tau}
\Ll^{\bet,\gamma,\tau}(\theta)
\;=\; \Ll_{\pred}(\theta)
\;+\; \lambda\, \Ll_{\ur}^{\bet,\gamma,\tau}(\theta),
\end{equation}
where $\Ll_{\ur}^{\bet,\gamma,\tau}$ is obtained from
\eqref{eq:L-beta-logtrace} by replacing the top-$n$ residual
$\sum_{j > n} \sigma_j^2$ with $\sum_{i \notin \mathcal{T}_\tau(x,r)}
\sigma_i^2$ in \eqref{eq:beta-hat}. Empirically on Galaxy10
(\S\ref{sec:beta-rehab-galaxy10}, Table~\ref{tab:galaxy10-matched}) the
eigenvalue-threshold rule adds no measurable peak-accuracy value
over the simpler fixed top-$n$ selection at $3$ seeds; the
conceptual advantages of (ii) carry through (local intrinsic-dim
estimate, robustness to misspecified $n$; cf.\
\S\ref{sec:eig-thresh-discussion}) but we recommend
$\Ll^{\bet,\gamma}$ as the practical default.

\subsubsection{Eigenvalue thresholding and local intrinsic dimensionality}\label{sec:eig-thresh-discussion}

The adaptive selection rule \eqref{eq:eig-thresh} introduced as the
second anti-collapse remedy is more than a numerical fix for the
point-mass degeneracy. It \emph{decouples} two roles that the
hyperparameter $n$ silently plays in the top-$n$ formulation:

\begin{enumerate}[leftmargin=*,topsep=2pt,itemsep=2pt]
\item \emph{Target rectifiability dimension.} The Carleson exponent
$r^{n+2}$ in \eqref{eq:beta-hat} and the dyadic weight $r^n$ in
\eqref{eq:L-beta} together encode the assumption ``$\mu$ is supported
on an $n$-dimensional set.'' This sets the \emph{geometric target}.

\item \emph{Tangent-subspace cardinality per anchor.} The choice to
sum the bottom $D - n$ eigenvalues of $S_r(x)$ in the residual
encodes ``locally, the cloud lies in an $n$-plane.'' This sets the
\emph{data-side description} of what counts as tangent at each
anchor.
\end{enumerate}

In the top-$n$ formulation these two roles are pinned to the same
integer. Real datasets, however, are rarely homogeneously
$n$-dimensional: the local intrinsic dimension typically varies
across $\supp \mu$ and across scales, since the underlying manifold
may pinch, branch, have boundary components, or contain regions of
different effective rank. The threshold rule replaces the rigid count
$n$ in role~(2) by a data-adaptive selection
$|\mathcal{T}_\tau(x, r)|$, while leaving the exponents in role~(1)
fixed at the global $n$. The selected count then doubles as a
\emph{local intrinsic-dimensionality estimate}: at each anchor $x$
and scale $r$, it reports how many independent directions of
variation are present, relative to the mean eigenvalue. This is
precisely the quantity that classical intrinsic-dimension estimators
\cite{FaccoTwoNN2017,AnsuiniID2019} return, but computed inline as a
free byproduct of the loss rather than via an auxiliary procedure.

\paragraph{The role of $\tau$.}
The threshold $\tau$ controls sensitivity to the local spectral gap.
We have three regimes:
\begin{itemize}[leftmargin=*,topsep=2pt,itemsep=2pt]
\item $\tau \ll 1$ (e.g.\ $\tau = 0.1$): includes near-tangent
directions with eigenvalues well below the mean. Acts as a soft
top-$n$ with a generous cutoff. Selected count is large.
\item $\tau \approx 1$: separates above-mean from below-mean
eigenvalues. On a rank-$r^*$ cloud with a clean spectral gap, this
recovers exactly the top-$r^*$ directions, regardless of the global
target $n$.
\item $\tau \gg 1$ (e.g.\ $\tau = 5$): restrictive; only sharply
peaked spectra survive. On an isotropic cloud, every eigenvalue
fails the cut and the residual equals the full trace, recovering an
isotropic-mass penalty.
\end{itemize}
$\tau$ is a one-dimensional hyperparameter that is in practice
cheaper to sweep than $n$ itself, since changes in $\tau$ shift the
implied local dim continuously rather than crossing the discrete
collapse threshold documented in \S\ref{sec:ablation-n}. A coarse grid
$\{0.5, 1, 2, 5\}$ covers the qualitative regimes; a finer sweep
around the winner is usually sufficient.

\paragraph{Relation to other intrinsic-dim approaches.}
Several SSL methods incorporate intrinsic-dimension awareness, but
through different mechanisms. MMCR \cite{MMCR2023} uses a global
nuclear-norm proxy that targets full-batch effective rank rather than
per-anchor local rank. VICReg's covariance term \cite{VICReg2022}
decorrelates global channel statistics without any local-rank
structure. Classical intrinsic-dim estimators
\cite{FaccoTwoNN2017,AnsuiniID2019} produce a single scalar per
dataset or per layer, decoupled from the training loss. The
threshold rule \eqref{eq:eig-thresh} differs from all three by
computing local intrinsic dimension at every anchor and every scale,
and by using the resulting count to weight the loss rather than
optimizing toward a prescribed dimension. The
\emph{geometric target} $n$ is given as input; the data is allowed
to declare its
\emph{local description} on its own terms.

\section{Practical considerations and failure modes}\label{sec:practical}

\subsection{Degenerate optima}\label{sec:degenerate-optima}
Every UR-style regularizer in this paper admits trivial minimizers,
just as LeJEPA's SIGReg can be zeroed by any Gaussian linear
reparameterization of the projector outputs. The predictive term
$\Ll_{\pred}$ of \eqref{eq:pred} provides the orthogonal
``invariance-plus-diversity'' signal that picks among the trivial set;
without $\Ll_{\pred}$, a constant $f_\theta$ would satisfy every
regularizer in the family. The interesting question is whether the
regularizer additionally admits a \emph{degenerate} minimum that
$\Ll_{\pred}$ alone cannot escape, because then training can be
trapped there. We address each loss in turn.

\paragraph{LeJEPA with $\Ll^{\text{SIGReg}}$ \cite{LeJEPA2025}.}
Zeroed exactly on the desired full-$D$ isotropic-Gaussian class. At
a point-mass cloud the 1-D marginals are $\delta$-distributions and
the Epps, Pulley characteristic-function statistic is bounded away
from zero, so $\Ll_{\pred}$ + SIGReg has no trivial point-mass
minimum.

\paragraph{UR--JEPA with $\Ll^{\text{CGLT}}$, dyadic-difference variant \eqref{eq:L-CGLT}.}
Zeroed when $\log \theta_r(x)$ is constant in $r$ at every anchor,
i.e.\ on the $n$-AD-regular UR class (the desired target). At
collapse, the Gaussian-smoothed density satisfies
$\theta_r(x) \propto r^{-n}$ near the cloud, so
$\Delta_r \log \theta = -n \log 2$ per dyadic scale; the loss takes
a fixed positive value at every point-mass configuration. No
trivial point-mass minimum.

\paragraph{UR--JEPA with $\Ll^{\text{CGLT},\partial\!\log}$, log-derivative variant \eqref{eq:L-CGLT-deriv}.}
Zeroed when $t \partial_t \log \theta_t = 0$ at every anchor and
scale, again the AD-regular UR class. At collapse,
$t \partial_t \log \theta_t = -n$, so the loss is bounded away from
zero by exactly the same mechanism as the dyadic-difference variant.

\paragraph{UR--JEPA with $\Ll^{\text{CGLT},\partial}$, raw scale-derivative variant \eqref{eq:L-CGLT-deriv-raw}.}
Zeroed on the smooth UR class characterized by
Theorem~\ref{thm:CGLT}\textnormal{(d)}. At collapse,
$\widetilde{\Delta}_{\mu_N,\varphi}(x, t) \sim -n / t^n$, which is
large in absolute value at every finite scale and diverges as the
smallest scale in the ladder tends to zero. No trivial point-mass
minimum.

\paragraph{UR--JEPA with $\beta$-loss.} $\Ll^\beta$ \eqref{eq:total-beta} is the one variant with a severe trivial minimum that
$\Ll_{\pred}$ alone cannot escape. Two distinct zero sets:
\begin{itemize}
\item[(i)] Exactly-$n$-flat configurations (the desired UR class):
the top-$n$ eigenvalues of $S_r(x)$ capture all the variance, so
the residual $\sum_{j > n} \sigma_j^2$ vanishes.
\item[(ii)] Point-mass configurations: $S_r(x) \equiv 0$, so every
eigenvalue vanishes and the residual vanishes trivially.
\end{itemize}
The second zero is a flat basin in loss space, and the $\beta$-loss
provides no gradient out of it. The two remedies of
\S\ref{sec:beta-anticollapse}, the intrinsic
$-\gamma \log \operatorname{tr}(S_r(x))$ penalty (yielding
$\Ll^{\bet,\gamma}$, \eqref{eq:total-beta-logtrace}) and
the adaptive eigenvalue-threshold selection (which, combined with
the log-trace penalty, yields $\Ll^{\bet,\gamma,\tau}$,
\eqref{eq:total-beta-gamma-tau}), are the design choices that break
this degeneracy in practice; under either remedy the point-mass
basin acquires a positive lower bound on the loss.

\paragraph{Summary.}
LeJEPA($\Ll^{\text{SIGReg}}$) and all three UR--JEPA($\Ll^{\text{CGLT},\,\star}$) variants are \emph{anti-collapse by
construction}: the regularizer alone bounds the loss away from zero
at any point-mass configuration. Only the bare $\beta$-number loss
has a trivial point-mass minimum and therefore requires an
additional remedy. The predictive loss $\Ll_{\pred}$ provides the
orthogonal ``select the UR structure consistent with augmentation
invariance'' signal in every case, but it does not by itself prevent
point-mass collapse.

\subsection{Target dimension $n$}
Unlike LeJEPA, UR--JEPA requires specifying an intrinsic dimension
$n$. Two options are used in this paper:
\begin{enumerate}[leftmargin=*]
\item \textbf{Fixed $n$} (default). Choose $n$ by domain knowledge or
a one-off estimate on a held-out run (e.g.\ the knee of the local-PCA
spectrum on a pre-trained encoder). The $n$-sweep of
\S\ref{sec:ablation-n} shows that probe accuracy is robust to the
exact value across a wide plateau.
\item \textbf{Threshold-adaptive local $n$} (for the $\beta$ variant;
\S\ref{sec:eig-thresh-discussion}). Fix the geometric target $n$
through the Carleson exponents, but let the per-anchor tangent count
vary via the threshold rule \eqref{eq:eig-thresh}. A scale-invariant
threshold $\tau$ is picked once; the local intrinsic dimension
$|\mathcal{T}_\tau(x,r)|$ adapts per anchor and per scale.
\end{enumerate}
Two alternative schedules (curriculum $n$, learned soft $n$) are
flagged in \S\ref{sec:future} as potential extensions but are not
evaluated here.

\subsection{Implementation notes}\label{sec:impl-notes}

\paragraph{Scale ladder.}
We set $r_{\max} = \mathrm{median}_{i}\,\|z_i - \bar z\|$ (a
diameter proxy) and $r_{\min} = r_{\max}\, N^{-1/n}$, with
$K = \lceil \log_{2}(r_{\max}/r_{\min}) \rceil$ dyadic levels. Both
endpoints are empirical and track the embedding distribution as
training evolves.

\paragraph{Minimum neighborhood count for $\beta$.}
For the $\beta$-loss to have a non-trivial gradient, $S_{r}(x)$
must have rank $\ge n + 1$; we therefore choose the smallest scale
$r_{K}$ large enough that $\sum_{j} w_{r_{K}}(z_{j} - x) \gtrsim n + 1$
on average.

\paragraph{Numerical scale across CGLT variants.}\label{sec:numerical-scale}
The three CGLT variants of \S\ref{sec:CGLTloss} (dyadic difference
\eqref{eq:L-CGLT}, log derivative \eqref{eq:L-CGLT-deriv}, and the
rescaled raw form \eqref{eq:L-CGLT-deriv-raw} via
\eqref{eq:Dtilde-rescaled}) all produce loss magnitudes in the same
$O(10)$ range on non-degenerate clouds, so the regularizer scaling
coefficient $s$ in
$\Ll = \lambda \cdot s \cdot \Ll_{\ur} + (1-\lambda)\Ll_{\pred}$
(\S\ref{sec:setup}) is shared across variants at $s = 10^{3}$.

\paragraph{Gradient noise.}
Because \eqref{eq:L-CGLT} is computed in log space via
\texttt{logsumexp}, contributions from far-away points are
exponentially suppressed in $\|z_{j} - x\|^{2}/r^{2}$ before the
log, so the gradient is dominated by neighbors within a few $r$ of
each anchor. The same exponential suppression applies to the
scale-derivative variants \eqref{eq:L-CGLT-deriv} and
\eqref{eq:L-CGLT-deriv-raw} since their per-scale quantity is a
Gaussian-kernel-weighted average. We use a row-wise exponent shift
in the log-derivative implementation for safety against underflow
at small $t$; the raw-derivative implementation uses the unshifted
kernel because the absolute magnitude of
$\widetilde{\Delta}$ enters the loss.

\paragraph{Distributed training.}
Both \eqref{eq:L-CGLT} and \eqref{eq:L-beta} are sums of local
statistics on anchors, with the neighbor set $Z$ living across
ranks. An all-gather of $Z$ makes the loss exact at $O(ND)$
communication; a chunked approximation that restricts each anchor's
neighbor set to its local shard preserves the estimator at $O(1)$
communication with a boundary bias. For LeJEPA-scale training runs,
the latter is preferred.

\section{Experiments}\label{sec:experiments}

We evaluate UR--JEPA against the LeJEPA baseline on ImageNet-10
(Inet10)~\cite{Imagenette2019,ImageNet2009},
ImageNet-100 (Inet100)~\cite{CMC2020,ImageNet2009}, and
Galaxy10~SDSS~\cite{Galaxy10,GalaxyZoo2008}. Because we experiment with multiple
uniform $n$-rectifiability losses for UR--JEPA, we denote by
UR--JEPA($\Ll$) the variant of UR--JEPA that uses the loss
$\Ll$, for
\[
\Ll \;\in\; \bigl\{\Ll^{\text{CGLT}},\;
\Ll^{\text{CGLT},\partial\!\log},\;
\Ll^{\text{CGLT},\partial},\;
\Ll^{\bet},\;
\Ll^{\bet,\gamma},\;
\Ll^{\bet,\gamma,\tau}\bigr\},
\]
corresponding respectively to the dyadic-difference Carleson loss
\eqref{eq:total-CGLT}, its log scale-derivative variant
\eqref{eq:L-CGLT-deriv}, its raw scale-derivative variant
\eqref{eq:L-CGLT-deriv-raw}, the $\beta$-number loss
\eqref{eq:total-beta}, and the latter augmented by the log-trace
penalty \eqref{eq:total-beta-logtrace} and additionally by the
adaptive eigenvalue-threshold rule \eqref{eq:total-beta-gamma-tau}.
Similarly, we denote by LeJEPA($\Ll^{\text{SIGReg}}$) the LeJEPA
baseline equipped with the SIGReg regularizer of \cite{LeJEPA2025}
defined in \eqref{eq:sigreg}.

\subsection{Setup}\label{sec:setup}

\paragraph{Architecture.}
All UR--JEPA experiments share a common pattern: a backbone
$f_\theta$ maps each augmented view to an embedding
$\mathrm{emb} \in \R^{d_{\mathrm{emb}}}$; a $3$-layer projector
$\mathrm{MLP}(d_{\mathrm{emb}}, [2048, 2048, D])$ with hidden-layer
\texttt{BatchNorm1d} (no norm on the output) produces the
$\mathrm{proj} \in \R^D$ vectors on which the regularizer acts; an
online linear probe
$\mathrm{LayerNorm}(d_{\mathrm{emb}}) \to \mathrm{Linear}(d_{\mathrm{emb}}, C)$
is trained jointly on detached $\mathrm{emb}$ to track downstream
accuracy without affecting the encoder. The backbone, the projector
output dimension $D$, and the number of classes $C$ depend on the
dataset.

\paragraph{Loss.}
\[
\Ll \;=\; \lambda \cdot s \cdot \Ll_{\mathrm{reg}}(\mathrm{proj})
\;+\; (1-\lambda)\cdot \Ll_{\mathrm{inv}}(\mathrm{proj}),
\qquad
\Ll_{\mathrm{inv}} = \|\overline{\mathrm{proj}} - \mathrm{proj}\|_2^2,
\]
with $\overline{\cdot}$ averaging across the $V$ augmented views.
For LeJEPA($\Ll^{\text{SIGReg}}$), $s = 1$ (its native scale). For all three UR--JEPA($\Ll^{\text{CGLT}}$)
variants (dyadic difference \eqref{eq:L-CGLT}, log derivative
\eqref{eq:L-CGLT-deriv}, and the $t_{\max}$-rescaled raw form
\eqref{eq:L-CGLT-deriv-raw} via \eqref{eq:Dtilde-rescaled}) and for
the $\beta$ family, $s = 10^3$ was chosen by a 5-point sweep so the
effective regularizer gradient matches LeJEPA($\Ll^{\text{SIGReg}}$)'s
$\lambda = 0.02$. The $t_{\max}$-rescaling in
\eqref{eq:Dtilde-rescaled} is what brings the raw form's magnitude
into the same $O(10)$ range as the log forms; see
\S\ref{sec:numerical-scale}. The AD anchor uses
$\lambda_{\mathrm{AD}} = 0.1$ for all CGLT variants.

\paragraph{Optimization.}
AdamW; weight decay $5 \times 10^{-2}$ on the backbone, $10^{-7}$
on the probe. Cosine schedule: one epoch of linear warmup from
$\mathrm{lr} / 100$, then cosine decay from $\mathrm{lr} = 2 \times
10^{-3}$ to $\eta_{\min} = 10^{-3}$ (following LeJEPA \texttt{MINIMAL.md}). 
\subsection{UR--JEPA vs.\ LeJEPA: Inet10}\label{sec:headline}

\begin{figure}[t]
\centering
\includegraphics[width=0.95\linewidth]{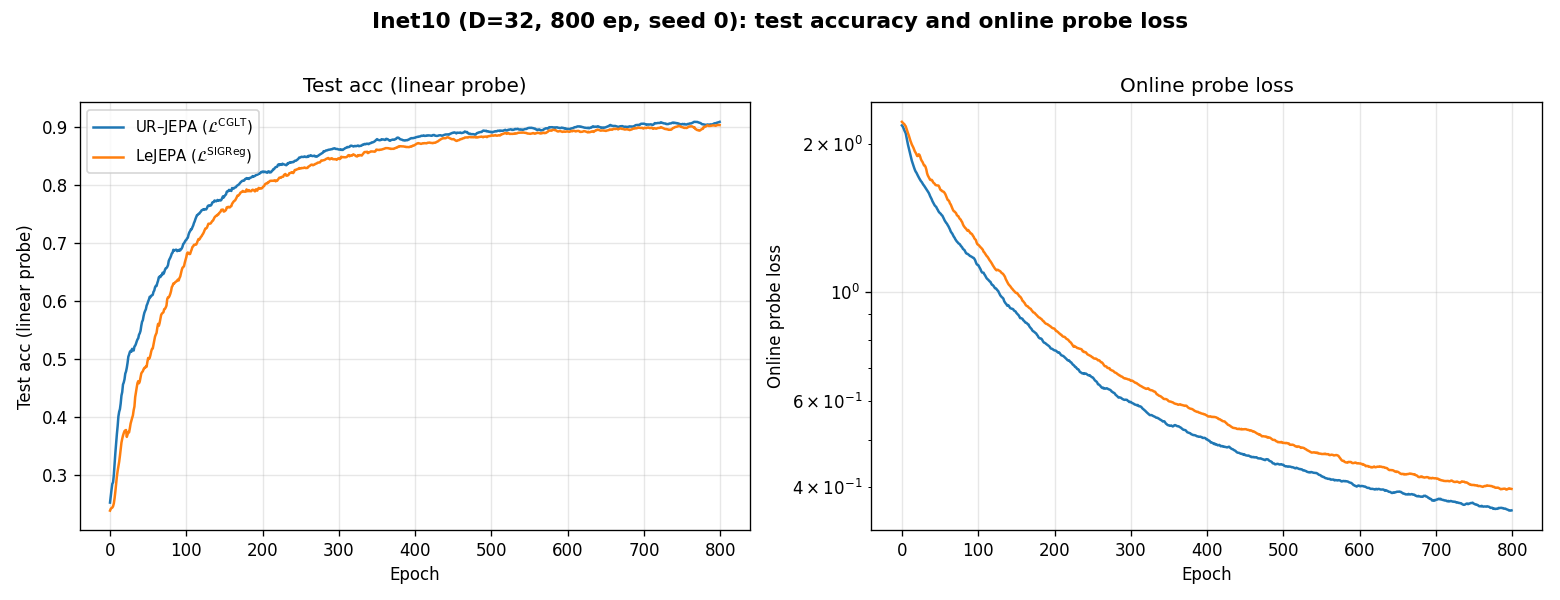}
\caption{Inet10 single-seed training at the headline configuration
($D = 32$, $n = 7$, $K = 5$): per-epoch linear-probe test accuracy
(left) and probe loss (right) across $800$ epochs for
LeJEPA($\Ll^{\text{SIGReg}}$) and UR--JEPA($\Ll^{\text{CGLT}}$). The figure
complements the peak-accuracy summary of Table~\ref{tab:headline}
by showing the full training dynamics underlying the $+0.83$\,pp
headline gap.}
\label{fig:inet10-trajectory}
\end{figure}

Table~\ref{tab:headline} reports best top-$1$ accuracy on the
Inet10 validation split across the $(\text{method}, D, n, K)$
configurations we screened. The headline result is at
$D = 32$, $n = 7$, $K = 5$: UR--JEPA($\Ll^{\text{CGLT}}$) attains
$0.9141 \pm 0.0014$, exceeding matched-recipe
LeJEPA($\Ll^{\text{SIGReg}}$) ($0.9058 \pm 0.0019$) by $+0.83$\,pp
(paired-$t = +15.5$, $p \ll 0.001$) with $\sim 30\%$ smaller seed
standard deviation. Figure~\ref{fig:inet10-trajectory} shows
UR--JEPA leading throughout training, with the gap established
early and held to convergence.

\begin{table}[h]
\centering
\small
\begin{tabular}{l c c c c c c}
\hline
method & $D$ & $n$ & $K$ & top-1 acc & seeds & wall (h)\\
\hline
LeJEPA($\Ll^{\text{SIGReg}}$) & $16$ & , & , &$0.9130 \pm 0.0008$ & $3$ & $\sim\!3.5$\\
LeJEPA($\Ll^{\text{SIGReg}}$) & $32$ & , & , &$0.9058 \pm 0.0019$ & $3$ & $\sim\!3.5$\\
UR--JEPA($\Ll^{\text{CGLT}}$) & $16$ & $8$ & $5$  & $0.9108 \pm 0.0004$ & $3$ & $\sim\!3.5$\\
UR--JEPA($\Ll^{\text{CGLT}}$) & $16$ & $7$ & $5$  & $0.9122 \pm 0.0019$ & $3$ & $\sim\!3.5$\\
UR--JEPA($\Ll^{\text{CGLT}}$) & $16$ & $7$ & $11$ & $0.9131 \pm 0.0013$ & $3$ & $\sim\!4.0$\\
UR--JEPA($\Ll^{\text{CGLT}}$) & $16$ & $7$ & $12$ & $0.9118 \pm 0.0018$ & $3$ & $\sim\!4.0$\\
\textbf{UR--JEPA($\Ll^{\text{CGLT}}$)} & $\mathbf{32}$ & $\mathbf{7}$ & $\mathbf{5}$
   & $\mathbf{0.9141 \pm 0.0014}$ & $3$ & $\sim\!3.5$\\
\cite{LeJEPA2025} published (best-of-sweep) & , & , & , & $0.907$ & , & ,\\
\hline
\end{tabular}
\caption{The LeJEPA published reference is a single-point best-of-sweep across
$\lambda$, $\mathrm{lr}$, and weight decay; our LeJEPA($\Ll^{\text{SIGReg}}$) numbers are
$3$-seed means at a single config and therefore expected to be slightly
below it. The bolded row is the overall best UR--JEPA($\Ll^{\text{CGLT}}$)
configuration. Wall hours are per seed on $1\times$ H100 at the
LeJEPA-exact recipe ($800$ epochs).}\label{tab:headline}
\end{table}

At $D = 16$ the two methods are statistically tied: tuning
UR--JEPA($\Ll^{\text{CGLT}}$) to $n = 7$ and extending the dyadic
ladder to $K = 11$ reaches $0.9131 \pm 0.0013$, matching
LeJEPA($\Ll^{\text{SIGReg}}$) at $0.9130 \pm 0.0008$.

\paragraph{Cross-$D$ asymmetry.}
Going from $D = 16$ to $D = 32$, LeJEPA($\Ll^{\text{SIGReg}}$)
degrades by $-0.72$\,pp while UR--JEPA($\Ll^{\text{CGLT}}$)
improves by $+0.19$\,pp; the SIGReg sketch at fixed
$\mathrm{num\_slices} = 256$ covers $S^{D-1}$ more sparsely as $D$
grows, whereas $\Ll^{\text{CGLT}}$'s local geometric statistics
carry no analogous coverage cost.

\subsubsection{Ablation: target intrinsic dimension $n$}\label{sec:ablation-n}

Table~\ref{tab:nsweep} sweeps the target intrinsic dimension at fixed
$D = 16$, and $K = 5$. The accuracy curve exhibits a sharp collapse
threshold at $n = 4$, and a wide plateau over
$n \in \{6, 7, 8, 9, 10\}$ with the peak at $n = 7$.

\begin{table}[h]
\centering
\small
\setlength{\tabcolsep}{5pt}
\begin{tabular}{c c c c}
\hline
$n$ & seeds & best acc & wall (h)\\
\hline
$4$           & $1$ & $0.2848$ \;\; (collapse)    & $\sim\!3.5$\\
$6$           & $3$ & $0.9055 \pm 0.0033$         & $\sim\!3.5$\\
$\mathbf{7}$  & $3$ & $\mathbf{0.9122 \pm 0.0019}$& $\sim\!3.5$\\
$8$           & $3$ & $0.9117 \pm 0.0009$         & $\sim\!3.5$\\
$9$           & $3$ & $0.9106 \pm 0.0013$         & $\sim\!3.5$\\
$10$          & $3$ & $0.9097 \pm 0.0024$         & $\sim\!3.5$\\
$12$          & $1$ & $0.9068$                    & $\sim\!3.5$\\
\hline
\end{tabular}
\caption{$n$-sweep at fixed $D = 16$, and $K = 5$ under the
LeJEPA-exact recipe ($400$ epochs; $3$ seeds per cell for
$n \in \{6, 7, 8, 9, 10\}$, $1$ seed for $n = 4$ and $n = 12$).
$n = 4$ collapses catastrophically (best acc $0.2848$). The peak
is at $n = 7$ ($0.9122 \pm 0.0019$), with $n = 8$
($0.9117 \pm 0.0009$) statistically indistinguishable. Probe
accuracy degrades by less than $1$\,pp across the plateau
$n \in \{6, 7, 8, 9, 10\}$ and by approximately $0.5$\,pp out to
$n = 12$. Wall hours are per seed on $1\times$ H100; the
regularizer cost is essentially independent of $n$.}\label{tab:nsweep}
\end{table}

The collapse threshold at $n = 4$ matches the theoretical analysis
of \S\ref{sec:practical}: at small $n$ the mismatch $(n - n^*)^2$ grows
quadratically and the optimizer breaks through to the regularizer's
degenerate fixed point (the constant). The wide plateau over
$n \in \{6, 7, 8, 9, 10\}$ indicates that
UR--JEPA($\Ll^{\text{CGLT}}$) does not require precise tuning
of $n$: the peak is at $n = 7$, with nearby values losing less than
a percentage point.

\subsubsection{Ablation: dyadic-scale count $K$}\label{sec:ablation-K}

Table~\ref{tab:ksweep} sweeps the number of dyadic scales $K$ at
fixed intrinsic dimension $n = 7$ (the $n$-sweep peak of
\S\ref{sec:ablation-n}) for both projector dimensions $D = 16$
and $D = 32$.

\begin{table}[h]
\centering
\small
\setlength{\tabcolsep}{5pt}
\begin{tabular}{c c c c c}
\hline
$D$ & $K$  & seeds & best acc & wall (h)\\
\hline
$16$ & $5$  & $3$ & $0.9122 \pm 0.0019$         & $\sim\!3.5$\\
$16$ & $11$ & $3$ & $\mathbf{0.9131 \pm 0.0013}$& $\sim\!4.0$\\
$16$ & $12$ & $3$ & $0.9118 \pm 0.0018$         & $\sim\!4.0$\\
\hline
$\mathbf{32}$ & $\mathbf{5}$ & $3$ & $\mathbf{0.9141 \pm 0.0014}$& $\sim\!3.5$\\
$32$ & $11$ & $3$ & $0.9116 \pm 0.0007$         & $\sim\!4.0$\\
$32$ & $12$ & $3$ & $0.9128 \pm 0.0012$         & $\sim\!4.0$\\
$32$ & $13$ & $3$ & $0.9113 \pm 0.0027$         & $\sim\!4.0$\\
$32$ & $14$ & $3$ & $0.9118 \pm 0.0017$         & $\sim\!4.0$\\
\hline
\end{tabular}
\caption{$K$-sweep on Inet10 at fixed $n = 7$ under the
LeJEPA-exact recipe, $3$ seeds per cell. Best acc is the maximum
online linear-probe top-1 accuracy across training epochs (mean
$\pm$ std over the $3$ seeds). Bold within each $D$ block marks
the maximum; the overall optimum is $D = 32$, $K = 5$ with
$0.9141 \pm 0.0014$. Wall hours are per seed on $1\times$ H100;
the regularizer's per-step cost grows mildly with $K$, so the
$K \ge 11$ cells run $\sim$0.5\,h longer than the $K = 5$ cells.}\label{tab:ksweep}
\end{table}

Two observations.
\textit{(i)} At $D = 16$, longer dyadic ladders provide a marginal
improvement: $K = 11$ ($0.9131 \pm 0.0013$) edges $K = 5$
($0.9122 \pm 0.0019$) by $+0.09$\,pp, well within combined seed
noise. Performance plateaus across $K \in \{5, 11, 12\}$.
\textit{(ii)} At $D = 32$, the $K = 5$ default is the overall
optimum and longer ladders ($K \in \{11, 12, 13, 14\}$) yield
slightly lower peak accuracy ($-0.13$\,pp to $-0.28$\,pp from the
peak), again within combined seed noise. Both observations
indicate that UR--JEPA($\Ll^{\text{CGLT}}$) is robust under
the choice of $K$ across an order-of-magnitude range from $K = 5$
to $K = 14$, and that $K = 5$ is an adequate default for the
recipes considered.

\subsection{UR--JEPA vs.\ LeJEPA: Galaxy10~SDSS}\label{sec:beta-rehab-galaxy10}

Galaxy10~SDSS~\cite{Galaxy10,GalaxyZoo2008} is a $10$-class
astronomical-morphology dataset ($\sim 21{,}800$ images at native
$69 \times 69$, upscaled to $128 \times 128$ via
\texttt{RandomResizedCrop}). We use the
\texttt{resnet18\_lowres} backbone of \S\ref{sec:setup} ($11$M
parameters) at $D = 32$, $n = 7$, $K = 5$, $\lambda = 0.02$,
$V = 4$, $\mathrm{bs} = 512$, $400$ epochs on a single H100, and
compare all UR--JEPA variants of \S\ref{sec:urjepa} against
LeJEPA($\Ll^{\text{SIGReg}}$) at this identical recipe.

\paragraph{Comparison to the published LeJEPA Table~5.}
At the matched-architecture $11$M-parameter point, our
$3$-seed UR--JEPA($\Ll^{\text{CGLT}}$) attains
$\mathbf{0.8142 \pm 0.0017}$, $+6.10$\,pp above LeJEPA's published
resnet18 cell ($0.7532$) and $+4.13$\,pp above LeJEPA's largest
cell (resnet34, $21$M, $2 \times$ our budget) at $0.7729$
(Table~\ref{tab:galaxy10-headline}). This gap, however, is not
purely regularizer-attributable: LeJEPA's published recipe differs
from ours in projector dimension, view count, image size, and
hyperparameters. Table~\ref{tab:galaxy10-matched} below isolates
the regularizer effect at our recipe.

\begin{table}[h]
\centering
\small
\setlength{\tabcolsep}{4pt}
\begin{tabular}{l l c c c c}
\hline
method & backbone & params & training & seeds & top-1 acc\\
\hline
IJEPA--IN22K \cite{IJEPA2023} & ViT-H/14         & 630M & IN22K found. & , & $0.6293$\\
LeJEPA \cite{LeJEPA2025}      & resnext26ts      & 8M   & $400$ ep     & , & $0.7378$\\
LeJEPA \cite{LeJEPA2025}      & swin\_tiny       & 27M  & $400$ ep     & , & $0.7489$\\
LeJEPA \cite{LeJEPA2025}      & resnet18         & 11M  & $400$ ep     & , & $0.7532$\\
LeJEPA \cite{LeJEPA2025}      & convnextv2\_nano & 14M  & $400$ ep     & , & $0.7605$\\
LeJEPA \cite{LeJEPA2025}      & resnet34         & 21M  & $400$ ep     & , & $0.7729$\\
\hline
\textbf{UR--JEPA($\Ll^{\text{CGLT}}$, ours)} & \texttt{resnet18\_lowres} & 11M & $400$ ep & 3 & $\mathbf{0.8142 \pm 0.0017}$\\
\hline
\end{tabular}
\caption{Galaxy10~SDSS published-comparison headline. Top block:
LeJEPA Table~5 \cite{LeJEPA2025} numbers for in-domain pretraining
across six backbones, plus the IJEPA--IN22K \cite{IJEPA2023}
foundation-model baseline reported there. Bottom row: our UR--JEPA($\Ll^{\text{CGLT}}$)
integral at the matched-architecture $11$M-parameter
point.}\label{tab:galaxy10-headline}
\end{table}

\begin{figure}[h]
\centering
\includegraphics[width=\textwidth]{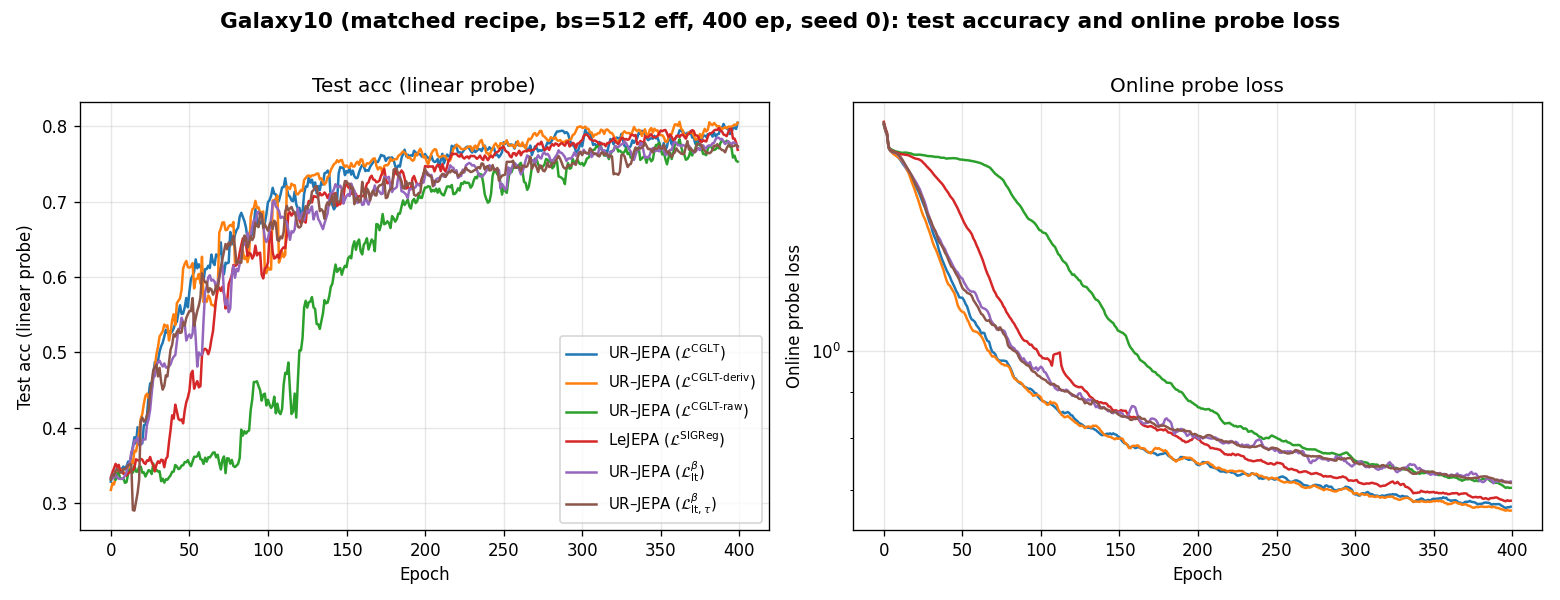}
\caption{Per-epoch online linear-probe top-1 accuracy (left) and
training-regularizer loss (right) trajectories on Galaxy10~SDSS,
for seed $0$ and $D=32$. Six variants are shown at the matched recipe:
UR--JEPA($\Ll^{\text{CGLT}}$), UR--JEPA
($\Ll^{\text{CGLT},\partial\!\log}$), UR--JEPA
($\Ll^{\text{CGLT},\partial}$), $\text{LeJEPA}(\Ll^{\text{SIGReg}})$, UR--JEPA
($\Ll^{\bet,\gamma}$), and UR--JEPA($\Ll^{\bet,\gamma,\tau}$). The
figure complements Table~\ref{tab:galaxy10-matched} by showing the
full training dynamics that
the peak-accuracy summary collapses to a single
number.}\label{fig:galaxy10-six-variant-acc-and-probe}
\end{figure}

\paragraph{Regularizer-only comparison.}
At our recipe, LeJEPA($\Ll^{\text{SIGReg}}$) attains
$0.8105 \pm 0.0050$ at $D = 32$, already $+5.73$\,pp above its
published resnet18 cell. UR--JEPA($\Ll^{\text{CGLT}}$) exceeds it
by $+0.37$\,pp on the point estimate ($t = 1.16$, $p \approx 0.37$
on $3$ matched-seed pairs), within seed noise. The $+6.10$\,pp gap
to LeJEPA-published therefore decomposes as $+5.73$\,pp from the
recipe and $+0.37$\,pp from the regularizer, with the latter not
significant at $3$ seeds. The per-epoch trajectories at seed $0$
for all six matched-recipe variants are plotted in
Figure~\ref{fig:galaxy10-six-variant-acc-and-probe}.

\begin{table}[h]
\centering
\footnotesize
\setlength{\tabcolsep}{4pt}
\begin{tabular}{l l c c c}
\hline
variant & basis & $D$ & best acc & wall (h)\\
\hline
\textbf{UR--JEPA($\Ll^{\text{CGLT}}$)}         & Thm.~\ref{thm:CGLT}(b), dyadic diff   & $32$ & $\mathbf{0.8142 \pm 0.0017}$ & $3.34$\\
UR--JEPA($\Ll^{\text{CGLT},\partial\!\log}$)   & Thm.~\ref{thm:CGLT}(d), log form      & $32$ & $0.8114 \pm 0.0037$         & $3.36$\\
LeJEPA($\Ll^{\text{SIGReg}}$) \cite{LeJEPA2025} & sliced char.\ fn.                    & $32$ & $0.8105 \pm 0.0050$         & $3.37$\\
LeJEPA($\Ll^{\text{SIGReg}}$) \cite{LeJEPA2025} & sliced char.\ fn.                    & $16$ & $0.8078 \pm 0.0017$         & $3.35$\\
UR--JEPA($\Ll^{\text{CGLT},\partial}$)         & Thm.~\ref{thm:CGLT}(d), raw Eq.~(1.5) & $32$ & $0.7963 \pm 0.0061$         & $3.23$\\
UR--JEPA($\Ll^{\bet,\gamma}$)                  & Jones $\beta$, log-trace remedy       & $32$ & $0.7956 \pm 0.0037$         & $6.07^{*}$\\
UR--JEPA($\Ll^{\bet,\gamma,\tau}$), $\tau=1.0$ & Jones $\beta$, log-trace + eig-thresh & $32$ & $0.7958 \pm 0.0068$         & $3.36$\\
\hline
\end{tabular}
\caption{Galaxy10~SDSS matched-recipe comparison
(\texttt{resnet18\_lowres}, $V = 4$, effective $\mathrm{bs} = 512$,
$400$~ep, $\lambda = 0.02$, $\lambda_{\ad} = 0.1$) sorted
by best top-1 acc. Wall hours are per seed on $1\times$ H100
unless marked: ${}^{*}$ UR--JEPA($\Ll^{\bet,\gamma}$) was run on
$2\times$ A100 DDP at $\mathrm{bs} = 256$/rank
(effective batch $512$, bit-identical to the single-GPU runs).}\label{tab:galaxy10-matched}
\end{table}

\paragraph{Findings.}
Four observations on Table~\ref{tab:galaxy10-matched}:
\textit{(i)} The three top variants
(UR--JEPA($\Ll^{\text{CGLT}}$),
UR--JEPA($\Ll^{\text{CGLT},\partial\!\log}$), and
LeJEPA($\Ll^{\text{SIGReg}}$) at $D = 32$) are statistically
indistinguishable, with the largest gap at $0.37$\,pp.
\textit{(ii)} UR--JEPA($\Ll^{\text{CGLT}}$) gives the headline:
highest mean ($0.8142$) and lowest variance ($\pm 0.0017$).
\textit{(iii)} UR--JEPA($\Ll^{\text{CGLT},\partial}$) (raw form)
underperforms UR--JEPA($\Ll^{\text{CGLT},\partial\!\log}$) by
$1.51$\,pp (over $2\sigma$), confirming that the log transform of
$\theta_t$ in \S\ref{sec:CGLTloss} is methodologically essential.
\textit{(iv)} LeJEPA's projector-dim preference is dataset-specific:
$-0.72$\,pp going $D = 16 \to 32$ on Inet10 but $+0.27$\,pp on
Galaxy10, so $D = 32$ is each method's optimum on this dataset.

\paragraph{The $\beta$ family.}
Both rehabilitated $\beta$ variants of \S\ref{sec:beta-anticollapse}
attain $\approx 0.7956\text{--}0.7958$,
$1.84\text{--}1.86$\,pp below the UR--JEPA($\Ll^{\text{CGLT}}$)
headline (paired-$t \approx 6$, $p \approx 0.02\text{--}0.03$,
significant). Adding the adaptive eigenvalue-threshold rule
(UR--JEPA($\Ll^{\bet,\gamma,\tau}$)) yields no peak-accuracy gain
over the log-trace penalty alone (UR--JEPA($\Ll^{\bet,\gamma}$)) at
$3$ seeds. The non-collapse result is nonetheless meaningful: bare
UR--JEPA($\Ll^{\bet}$) is known to collapse on Inet10 across $5$
orders of magnitude of $s$, but pairing it with
the log-trace penalty restores competitive performance.

\subsection{UR--JEPA vs.\ LeJEPA: Inet100}\label{sec:inet100}

We extend the matched-recipe comparison to
Inet100~\cite{CMC2020} ($100$ classes,
$\sim 130{,}000$ training images), $10\times$ the Inet10
training-set size of \S\ref{sec:headline}. The recipe is
the same configuration of \S\ref{sec:setup}
(\texttt{vit\_small\_patch8\_224} backbone at input size $128$,
$V = 4$ augmented views, $D = 32$, $n = 7$, $K = 5$,
$\lambda = 0.02$, $s = 10^3$, batch size $256$,
$\mathrm{lr} = 2 \times 10^{-3}$ cosine-decayed to
$\mathrm{lr}_{\min} = 10^{-3}$, run on one H100, trained for
$400$ epochs and resumed to $800$ epochs).

\begin{figure}[t]
\centering
\includegraphics[width=0.95\linewidth]{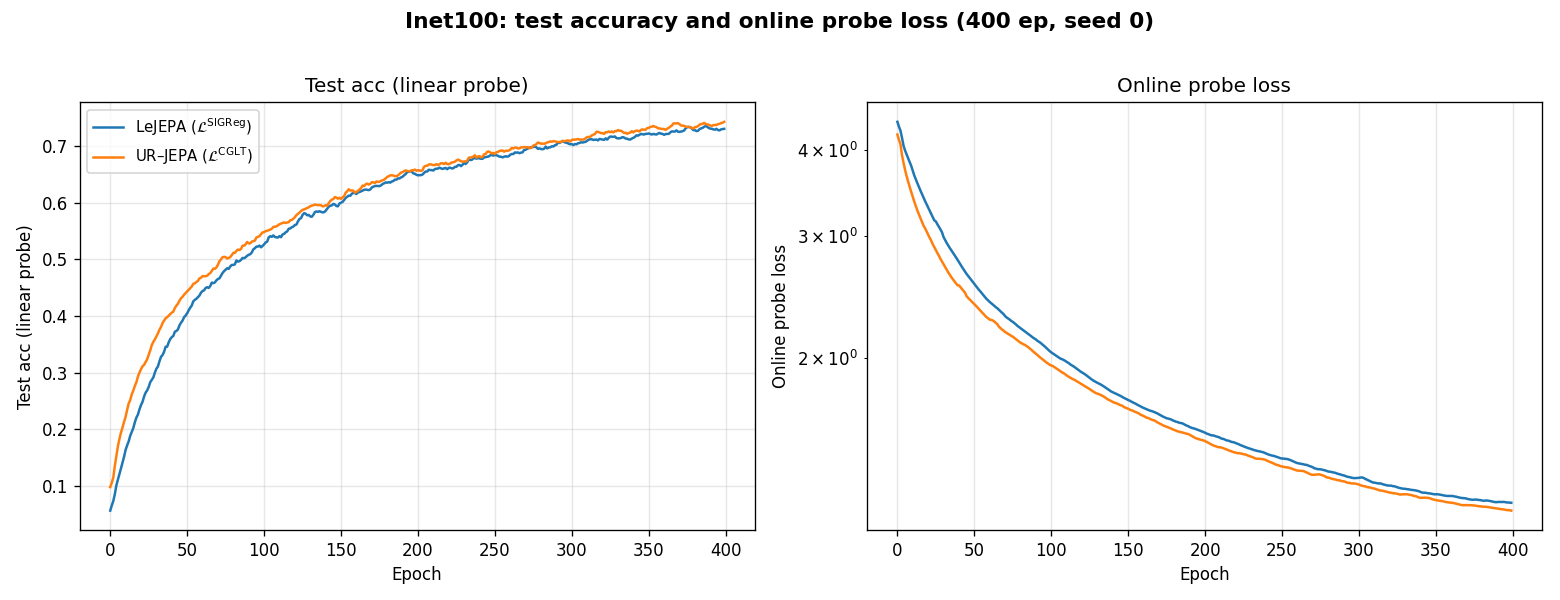}
\caption{Inet100 single-seed training: linear-probe test
accuracy (left) and probe loss (right) across $400$ epochs for
LeJEPA($\Ll^{\text{SIGReg}}$) and UR--JEPA($\Ll^{\text{CGLT}}$). The
accuracy gap, large at epoch $100$, contracts steadily across the
shown range; extended training to $800$ epochs (Table~\ref{tab:inet100-trajectory})
reopens the gap to $+1.58$\,pp.}
\label{fig:inet100-trajectory}
\end{figure}

\paragraph{Convergence trajectory.}
Table~\ref{tab:inet100-trajectory} shows the probe accuracy
trajectory. Both methods gain $\sim 19$\,pp going from epoch $100$
to epoch $400$, while the UR--JEPA$-$LeJEPA difference contracts
from $+2.02$\,pp to $+0.62$\,pp. Continued training to epoch $800$
adds another $\sim 2.5$\,pp to each method and the gap re-opens to
$+1.58$\,pp, the largest in-domain Inet100 advantage we observe at
this checkpoint. The peak-accuracy gap at $800$\,ep is in the same
small-effect band as the matched-recipe Galaxy10 finding
($+0.37$\,pp at $3$ seeds, paired-$t = 1.16$, $p \approx 0.37$),
amplified by roughly $4\times$ at the longer Inet100 schedule. The
per-epoch test-accuracy and probe-loss curves are plotted in
Figure~\ref{fig:inet100-trajectory}.

\begin{table}[h]
\centering
\small
\setlength{\tabcolsep}{5pt}
\begin{tabular}{c c c c}
\hline
epoch & LeJEPA($\Ll^{\text{SIGReg}}$) & UR--JEPA($\Ll^{\text{CGLT}}$) & $\Delta$\\
\hline
$100$ & $0.5290$ & $0.5492$ & $+2.02$\,pp\\
$200$ & $0.6406$ & $0.6598$ & $+1.92$\,pp\\
$300$ & $0.7000$ & $0.7084$ & $+0.84$\,pp\\
$400$ & $0.7402$ & $0.7464$ & $+0.62$\,pp\\
$800$ & $0.7590$ & $\mathbf{0.7748}$ & $+1.58$\,pp\\
\hline
\end{tabular}
\caption{Inet100 single-seed convergence trajectory. Online
linear-probe top-$1$ accuracy at seed $0$ for
LeJEPA($\Ll^{\text{SIGReg}}$) and UR--JEPA($\Ll^{\text{CGLT}}$)
measured at epochs $\{100, 200, 300, 400, 800\}$. The column
$\Delta = \text{UR--JEPA} - \text{LeJEPA}$ is non-monotone: it
contracts from $+2.02$\,pp at partial training to $+0.62$\,pp at
$400$ epochs, then re-opens to $+1.58$\,pp at $800$ epochs.}
\label{tab:inet100-trajectory}
\end{table}

\paragraph{Downstream transfer.}
We evaluate frozen-backbone linear-probe transfer from the
$800$-epoch Inet100 checkpoints onto five LeJEPA Table~$5$
datasets (Aircraft, CIFAR-$100$, DTD, Flowers, Food), following
LeJEPA's published probe protocol (features = concat of the last
two CLS tokens, $50$-epoch AdamW probe with cosine annealing to
$10^{-6}$, single seed per cell).
Table~\ref{tab:inet100-transfer} summarizes the results: the mean
transfer $\Delta$ across the five datasets is $+0.32$\,pp;
UR--JEPA leads on $4$ of $5$ datasets (Aircraft, DTD, Flowers,
Food) and trails on CIFAR-$100$, with the largest single-dataset
gap at $+1.11$\,pp on Aircraft.

\begin{table}[h]
\centering
\small
\setlength{\tabcolsep}{5pt}
\begin{tabular}{l c c c c c c}
\hline
pretrain & Aircraft & CIFAR-100 & DTD & Flowers & Food & row mean\\
\hline
LeJEPA($\Ll^{\text{SIGReg}}$)   & $0.3654$ & $\mathbf{0.6533}$ & $0.5867$ & $0.7676$ & $0.6130$ & $0.5972$\\
UR--JEPA($\Ll^{\text{CGLT}}$) & $\mathbf{0.3765}$ & $0.6448$ & $\mathbf{0.5941}$ & $\mathbf{0.7699}$ & $\mathbf{0.6167}$ & $\mathbf{0.6004}$\\
\hline
$\Delta$ (UR--JEPA$-$LeJEPA) & $+0.0111$ & $-0.0085$ & $+0.0074$ & $+0.0023$ & $+0.0037$ & $+0.0032$\\
\hline
\end{tabular}
\caption{Downstream linear-probe transfer from the $800$-epoch
Inet100 seed-$0$ checkpoints to five LeJEPA Table~$5$ datasets.
Each cell is the best top-$1$ test accuracy over a $50$-epoch
linear probe on frozen backbone features. The mean transfer
$\Delta = +0.32$\,pp across the five datasets, with UR--JEPA
leading on $4$ of $5$; the largest single-dataset gap is
$+1.11$\,pp on Aircraft and the only adverse cell is
CIFAR-$100$ at $-0.85$\,pp. Single seed per cell; three-seed
verification is deferred.}
\label{tab:inet100-transfer}
\end{table}

\paragraph{Epoch ablation on transfer.}
Probing the same backbones at the earlier $400$-epoch checkpoint
yields a mean transfer $\Delta$ of $+0.27$\,pp (UR--JEPA ahead on
$3$ of $5$), so extending pretraining from $400$ to $800$ epochs
nudges the head-to-head only marginally ($+0.27 \to +0.32$\,pp)
while lifting both methods by $\sim 3.2$\,pp in mean transfer
accuracy (LeJEPA $+3.16$\,pp, UR--JEPA $+3.21$\,pp). The extra
training therefore benefits both regularizers approximately
equally and the per-dataset $\Delta$ values are jittery at one
seed, reinforcing the directional rather than significance-backed
reading of the matrix above.

\begin{remark}
At the matched recipe and one seed per method,
UR--JEPA($\Ll^{\text{CGLT}}$) is sample-efficient
(Table~\ref{tab:inet100-trajectory} and
Figure~\ref{fig:inet100-trajectory}: a $+2.02$\,pp lead at epoch
$100$ that contracts to $+0.62$\,pp at epoch $400$ then re-opens
to $+1.58$\,pp at epoch $800$) and ahead on $4$ of $5$ downstream
transfer datasets at the $800$-epoch checkpoint
(mean $\Delta = +0.32$\,pp; Table~\ref{tab:inet100-transfer}). The
single-seed evidence on Inet100 is therefore directionally
consistent with the matched-recipe Galaxy10 picture
(\S\ref{sec:beta-rehab-galaxy10}) but at a larger effect size; a
$3$-seed promotion is required before any
regularizer-attributable claim on Inet100 is statistically
backed.
\end{remark}

\subsection{UR--JEPA vs.\ LeJEPA: EuroSAT (remote sensing)}\label{sec:eurosat}

We add a second non-natural-image domain alongside Galaxy10
to test whether the matched-recipe picture generalizes:
EuroSAT~RGB~\cite{Helber2019}, a $10$-class Sentinel-$2$
land-cover dataset ($16{,}200$ train / $5{,}400$ val /
$5{,}400$ test, native $64 \times 64$). The recipe is the
Galaxy10 matched configuration of
\S\ref{sec:beta-rehab-galaxy10}
(\texttt{resnet18\_lowres}, $D = 32$, $V = 4$, $n = 7$,
$K = 5$, $\lambda = 0.02$, $s = 10^{3}$,
$\lambda_{\ad} = 0.1$, $400$ epochs, $V = 4$), the only
deviation being effective batch size $256$ on a single
A100-$40$\,GB (the no-maxpool low-resolution path
out-of-memories at $\mathrm{bs} = 512$). Both
LeJEPA($\Ll^{\text{SIGReg}}$) and UR--JEPA($\Ll^{\text{CGLT}}$)
are run at $3$ seeds.

\paragraph{Probe accuracy: statistically tied, UR--JEPA $\sim\!2\times$ tighter variance.}
Best online linear-probe top-$1$ accuracy across training:

\begin{table}[h]
\centering
\small
\setlength{\tabcolsep}{5pt}
\begin{tabular}{l c c}
\hline
method & best acc (mean $\pm$ std, $3$ seeds) & wall (h) \\
\hline
LeJEPA($\Ll^{\text{SIGReg}}$)   & $\mathbf{0.9611 \pm 0.0019}$ & $\sim\!3.4$ \\
UR--JEPA($\Ll^{\text{CGLT}}$) & $0.9597 \pm 0.0009$         & $\sim\!3.4$ \\
\hline
$\Delta$ (UR--JEPA $-$ LeJEPA) & $-0.0014$ & , \\
\hline
\end{tabular}
\caption{EuroSAT~RGB matched-recipe head-to-head, $3$ seeds
per cell. Per-seed paired $\Delta = \{0.0000, -0.0018,
-0.0023\}$, paired-$t \approx -1.96$ on $2$~dof, two-tailed
$p \approx 0.19$: the two methods are statistically tied with
LeJEPA nominally ahead. Wall hours are per seed on $1\times$
A100-$40$\,GB at the matched recipe (effective batch $256$).}
\label{tab:eurosat}
\end{table}

The accuracy gap of $-0.14$\,pp is within seed noise (paired
$t \approx -1.96$, $p \approx 0.19$), the same statistical
tie observed at convergence on Galaxy10 ($+0.37$\,pp,
$p \approx 0.37$; \S\ref{sec:beta-rehab-galaxy10}) and on
Inet100 ($+1.58$\,pp at one seed; \S\ref{sec:inet100}). The
UR--JEPA seed standard deviation ($\pm 0.0009$) is roughly
half that of LeJEPA ($\pm 0.0019$), continuing the
lower-variance pattern from Inet10
($\pm 0.0004$ vs $\pm 0.0008$; \S\ref{sec:headline}) and
Galaxy10 ($\pm 0.0017$ vs $\pm 0.0050$;
\S\ref{sec:beta-rehab-galaxy10}). EuroSAT is therefore the
third dataset confirming that UR--JEPA delivers lower seed
variance than LeJEPA at the matched recipe.

\paragraph{Foundation-model comparison.}
Table~\ref{tab:eurosat-foundation} sets our $11$M-parameter
\texttt{resnet18\_lowres} cells ($96.0$ to $96.1\%$) against
the frozen-feature transfer baselines collected by
\cite{CorleyEuroSATBenchmark2024}. In-domain UR--JEPA and
LeJEPA exceed ImageNet-supervised transfer ($82$ to $93\%$)
and the same-scale remote-sensing foundation models GASSL and
SeCo at native $64$\,px ($89.5$ to $93.1\%$), and match
Scale-MAE at the ViT-L scale ($96.00\%$) with a $25\times$
smaller backbone. They do not match the largest $224$\,px
remote-sensing foundation models (SatMAE ViT-L $98.94\%$,
SeCo at $224$\,px $96.30\%$). This shows that \emph{in-domain SSL at $11$M parameters is
competitive with large foundation-model transfer on remote
sensing, matching or exceeding same-scale and smaller-scale
baselines}, in contrast with the larger margin observed on
Galaxy10 (\S\ref{sec:beta-rehab-galaxy10}). 
\begin{table}[h]
\centering
\footnotesize
\setlength{\tabcolsep}{4pt}
\begin{tabular}{l l r l c c}
\hline
model & backbone & params & pretrain & img & EuroSAT \\
\hline
ImageNet-sup    & ResNet-$50$               & $25$M           & ImageNet               & $64$            & $82.09$/$86.44$ \\
GASSL           & ResNet-$18$               & $11$M           & RS foundation          & $64$            & $89.51$ \\
SeCo            & ResNet-$18$               & $11$M           & RS foundation          & $64$            & $93.14$ \\
ImageNet-sup    & ResNet-$50$               & $25$M           & ImageNet               & $224$           & $93.13$ \\
MoCo-ImageNet   & ResNet-$50$               & $25$M           & ImageNet SSL           & $64$            & $94.11$ \\
MoCo-ImageNet   & ResNet-$50$               & $25$M           & ImageNet SSL           & $224$           & $95.76$ \\
\textbf{UR--JEPA($\Ll^{\text{CGLT}}$) (ours)} & \texttt{resnet18\_lowres} & $11$M & \textbf{in-domain SSL} & $64\!\to\!128$ & $\mathbf{95.97 \pm 0.09}$ \\
Scale-MAE       & ViT-L                     & $\sim\!300$M    & RS foundation          & $64$            & $96.00$ \\
\textbf{LeJEPA($\Ll^{\text{SIGReg}}$) (ours)} & \texttt{resnet18\_lowres} & $11$M & \textbf{in-domain SSL} & $64\!\to\!128$ & $\mathbf{96.11 \pm 0.19}$ \\
SeCo            & ResNet-$18$               & $11$M           & RS foundation          & $224$           & $96.30$ \\
SatMAE          & ViT-L                     & $\sim\!300$M    & RS foundation          & $224$           & $98.94$ \\
\hline
\end{tabular}
\caption{EuroSAT RGB benchmark baselines from
\cite{CorleyEuroSATBenchmark2024}, sorted by accuracy. All
rows except the bolded ones are \emph{frozen-feature transfer}
(linear-probe accuracy shown where reported; KNN-$5$
otherwise) from external pretraining. The bolded rows are our
\emph{in-domain} matched-recipe
LeJEPA($\Ll^{\text{SIGReg}}$) and UR--JEPA($\Ll^{\text{CGLT}}$)
cells of Table~\ref{tab:eurosat}, evaluated by linear probe on
frozen in-domain features and reported as mean $\pm$ std over
$3$ seeds. Numbers across rows are not strictly matched; see
caveats in the prose.}
\label{tab:eurosat-foundation}
\end{table}

\begin{remark}
At the matched recipe and $3$ seeds per method on EuroSAT,
UR--JEPA($\Ll^{\text{CGLT}}$) and
LeJEPA($\Ll^{\text{SIGReg}}$) are statistically tied on
peak probe accuracy (paired $t \approx -1.96$,
$p \approx 0.19$, $\Delta = -0.14$\,pp), with
UR--JEPA holding $\sim\!2\times$ tighter seed variance.
Together with the matched-recipe results on Inet10
(\S\ref{sec:headline}), Galaxy10
(\S\ref{sec:beta-rehab-galaxy10}), and Inet100
(\S\ref{sec:inet100}), EuroSAT becomes the fourth dataset
in the same accuracy-band picture; the geometrically
distinct projector signature is documented at $3$ seeds in
\S\ref{sec:viz-geometry} (Figure~\ref{fig:viz-geometry-eurosat-6way},
Table~\ref{tab:viz-eurosat}).
\end{remark}

\subsection{Geometric distinction between LeJEPA and UR--JEPA}\label{sec:viz-geometry}

The matched-recipe comparisons in \S\ref{sec:beta-rehab-galaxy10}
and \S\ref{sec:inet100} place LeJEPA($\Ll^{\text{SIGReg}}$) and
UR--JEPA($\Ll^{\text{CGLT}}$) in the same peak-accuracy band at
convergence. We now ask whether this accuracy parity reflects
representational parity, by directly inspecting the distribution of
projector outputs under each regularizer.

\paragraph{Setup.}
For each of Inet10, Galaxy10~SDSS, and Inet100 we load the
seed-$0$ matched-recipe checkpoint of LeJEPA($\Ll^{\text{SIGReg}}$)
and UR--JEPA($\Ll^{\text{CGLT}}$) (projector dimension $D = 32$),
pass the validation split through the trained backbone and
projector at $V = 1$ test transform, and collect
$\sim 5000$ projector outputs $Z \in \mathbb{R}^{N \times D}$.
On $Z$ we then compute (a) the eigenvalue spectrum of the
empirical covariance $Z_c^\top Z_c / (N - 1)$ and (b) the
per-dimension Shapiro-Wilk $W$ statistic, a goodness-of-fit
measure for marginal Gaussianity ($W = 1$ for an exact Gaussian).

\begin{figure}[t]
\centering
\includegraphics[width=0.95\linewidth]{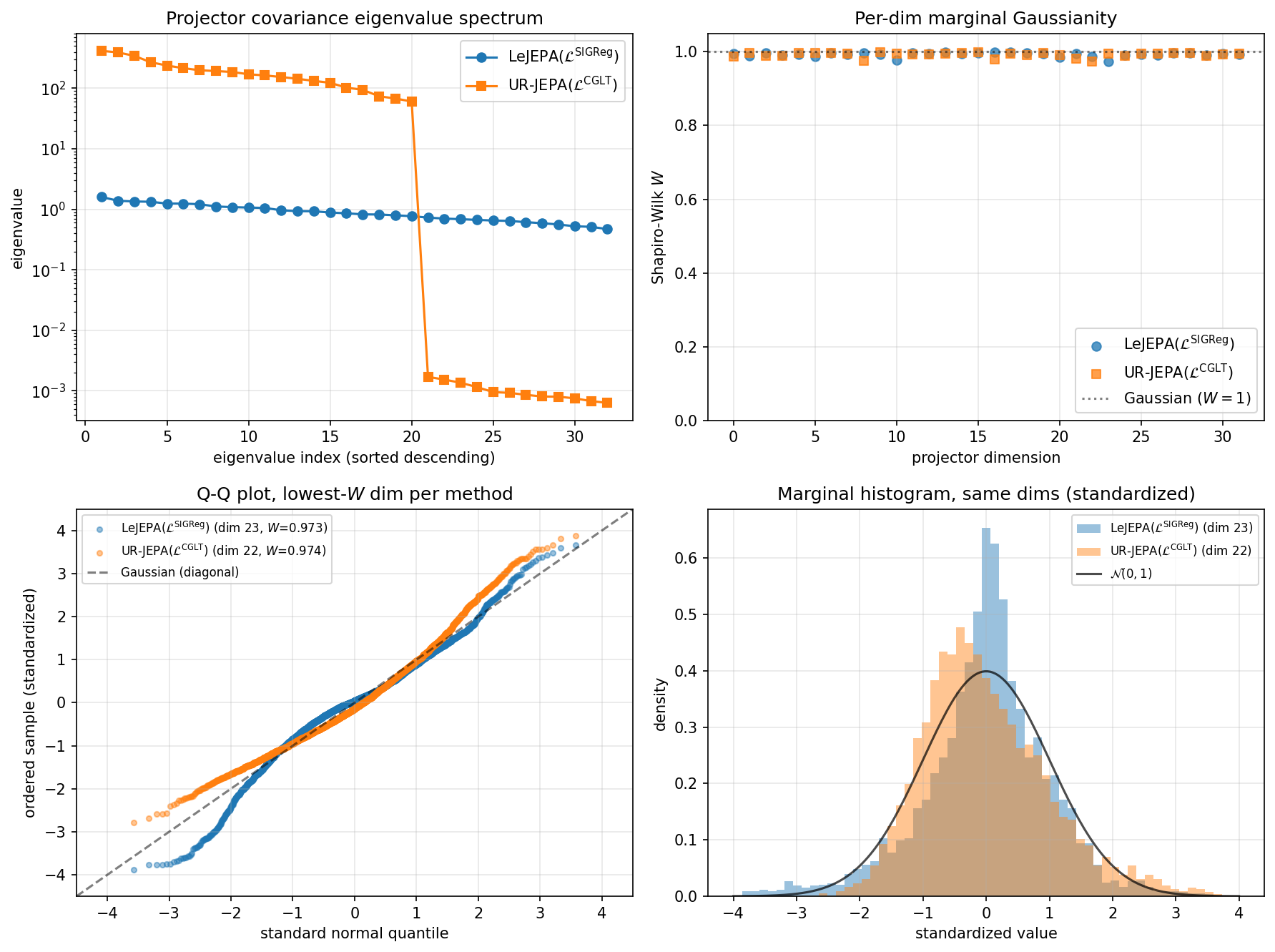}
\caption{Projector geometry diagnostics on Inet10 at the seed-$0$
matched-recipe checkpoints. Top-left: covariance eigenvalue spectrum
on log-y; LeJEPA($\Ll^{\text{SIGReg}}$) is near-isotropic
(top-to-bottom ratio $3.39$), UR--JEPA($\Ll^{\text{CGLT}}$) shows a
sharp drop (ratio $\sim 6.6 \times 10^5$). Top-right: per-dimension
Shapiro-Wilk $W$; both methods cluster near $W = 1$. Bottom-left:
Q-Q plot at each method's lowest-$W$ dimension. Bottom-right:
standardized marginal histogram at the same dimensions with
$\mathcal{N}(0,1)$ overlay.}
\label{fig:viz-geometry-inet10}
\end{figure}

\begin{figure}[t]
\centering
\includegraphics[width=0.95\linewidth]{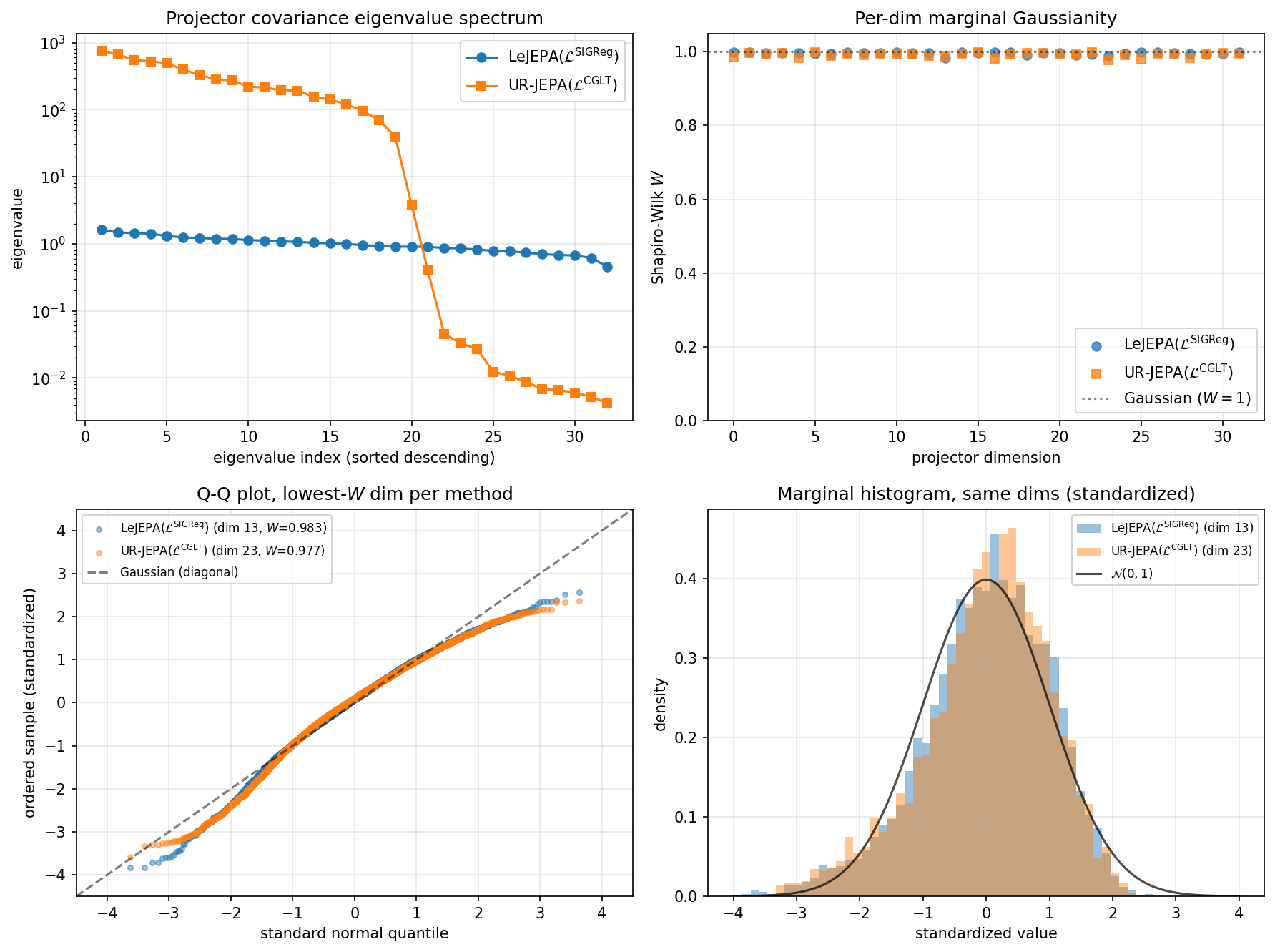}
\caption{Projector geometry diagnostics on Galaxy10~SDSS at the
seed-$0$ matched-recipe checkpoints. Top-left: covariance eigenvalue
spectrum on log-y; LeJEPA($\Ll^{\text{SIGReg}}$) is near-isotropic
(top-to-bottom ratio $3.57$), UR--JEPA($\Ll^{\text{CGLT}}$) shows a
sharp drop (ratio $\sim 1.8 \times 10^5$). The marginal-Gaussianity
parity from Inet10 replicates here.}
\label{fig:viz-geometry-galaxy10}
\end{figure}

\begin{figure}[t]
\centering
\includegraphics[width=0.95\linewidth]{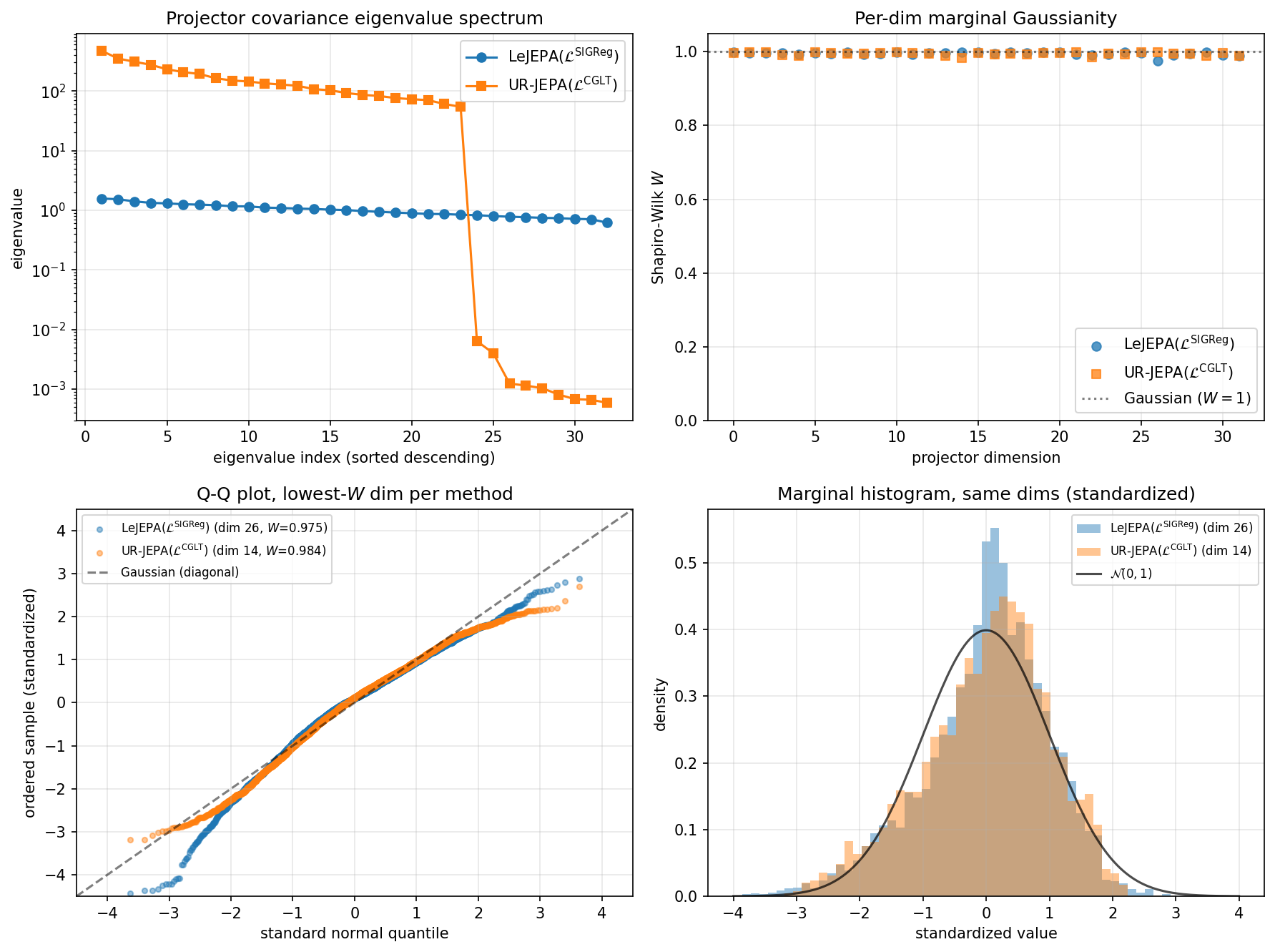}
\caption{Projector geometry diagnostics on the Inet100
validation split at the seed-$0$ matched-recipe checkpoints (the
same checkpoints whose trajectories are plotted in
Figure~\ref{fig:inet100-trajectory}). Top-left: covariance
eigenvalue spectrum on log-y; LeJEPA($\Ll^{\text{SIGReg}}$) is
near-isotropic (top-to-bottom ratio $2.54$),
UR--JEPA($\Ll^{\text{CGLT}}$) shows a sharp drop at index $\sim 25$
with top-to-bottom ratio $\sim 8 \times 10^5$, the largest of the
three datasets.}
\label{fig:viz-geometry-inet100}
\end{figure}

Figures~\ref{fig:viz-geometry-inet10},
\ref{fig:viz-geometry-galaxy10}, \ref{fig:viz-geometry-inet100},
and \ref{fig:viz-geometry-eurosat-6way} report four diagnostics
computed from the projector outputs collected over the validation
split of each of Inet10, Galaxy10~SDSS, Inet100, and EuroSAT, at
the matched-recipe checkpoints of LeJEPA($\Ll^{\text{SIGReg}}$)
and UR--JEPA($\Ll^{\text{CGLT}}$). The first three figures show
seed-$0$ projector outputs; the EuroSAT figure overlays all $3$
seeds of each method (six traces). The four figures share the
same $2 \times 2$ layout.
The \emph{top-left} panel plots the covariance eigenvalue spectrum
on log-y, sorted in descending order; a near-flat curve indicates
isotropy (equal variance in every direction), while a sharply
declining curve indicates an effectively low-rank distribution
concentrated on a few directions. The \emph{top-right} panel
reports the per-dimension Shapiro-Wilk $W$ statistic, a
goodness-of-fit measure for marginal Gaussianity: $W = 1$
corresponds to an exact Gaussian, $W < 0.95$ signals visible
non-Gaussianity. The \emph{bottom-left} panel is a Q-Q plot at
each method's lowest-$W$ dimension (the most non-Gaussian marginal
of each method), where points along the dashed diagonal indicate
Gaussianity and S-shape deviation indicates heavier-than-Gaussian
tails. The \emph{bottom-right} panel overlays the standardized
marginal histogram at the same dimensions on $\mathcal{N}(0,1)$;
bulk-shape disagreement shows up as asymmetry or multimodality.

\paragraph{Eigenvalue spectrum.}
Across all three datasets, LeJEPA($\Ll^{\text{SIGReg}}$) yields a
near-flat spectrum (top-to-bottom eigenvalue ratio at most $3.6$),
consistent with finite-sample fluctuation around an isotropic
Gaussian. UR--JEPA($\Ll^{\text{CGLT}}$) yields a sharply two-regime
spectrum: roughly $20$ to $25$ active dimensions carrying
eigenvalues from $10^0$ to $10^3$, followed by a $4$ to $5$
order-of-magnitude descent to $\sim 10^{-3}$ for the remaining
dimensions. The effect is consistent across all three datasets in
absolute and relative terms (Table~\ref{tab:viz-eig-ratio},
Figures~\ref{fig:viz-geometry-inet10},
\ref{fig:viz-geometry-galaxy10},
\ref{fig:viz-geometry-inet100}).

\begin{table}[h]
\centering
\small
\setlength{\tabcolsep}{5pt}
\begin{tabular}{l c c c}
\hline
dataset & LeJEPA($\Ll^{\text{SIGReg}}$) & UR--JEPA($\Ll^{\text{CGLT}}$) & ratio (UR/LeJEPA)\\
\hline
Inet10            & $3.39$ & $657{,}111$ & $\sim 1.9 \times 10^{5}$ \\
Galaxy10~SDSS     & $3.57$ & $177{,}555$ & $\sim 5 \times 10^{4}$   \\
Inet100      & $2.54$ & $798{,}642$ & $\sim 3.1 \times 10^{5}$ \\
\hline
\end{tabular}
\caption{Eigenvalue ratio $\mathrm{eig}[0] / \mathrm{eig}[-1]$ of
the projector covariance at the seed-$0$ matched-recipe checkpoint
of each method. LeJEPA's sliced-characteristic-function loss drives
the projector toward isotropy; UR--JEPA's $\Ll^{\text{CGLT}}$ leaves
the global covariance with a $4$ to $5$ order-of-magnitude
effective-rank reduction.}
\label{tab:viz-eig-ratio}
\end{table}


\paragraph{Marginal Gaussianity and the Diaconis-Freedman
phenomenon.}
Both methods produce per-dimension marginals close to
$\mathcal{N}(0, 1)$ (mean Shapiro-Wilk $W$~\cite{ShapiroWilk1965}
$= 0.992$ to $0.996$ across all three datasets), despite only
LeJEPA($\Ll^{\text{SIGReg}}$) explicitly targeting Gaussianity
through its sliced characteristic-function objective. This is
consistent with the Diaconis-Freedman
phenomenon~\cite{DiaconisFreedman1984}: typical low-dimensional
linear projections of a high-dimensional measure are approximately
Gaussian under mild regularity, so projector outputs in
$\mathbb{R}^{32}$ satisfy the marginal-Gaussianity property without
it being explicitly enforced. The worst dimension of each method
still admits a Q-Q plot within mild deviation of $\mathcal{N}(0,1)$
(bottom panels of Figures~\ref{fig:viz-geometry-inet10},
\ref{fig:viz-geometry-galaxy10}, and
\ref{fig:viz-geometry-inet100}). The two methods diverge in the
covariance structure, not in per-dimension distribution shape.

\begin{figure}[t]
\centering
\includegraphics[width=0.95\linewidth]{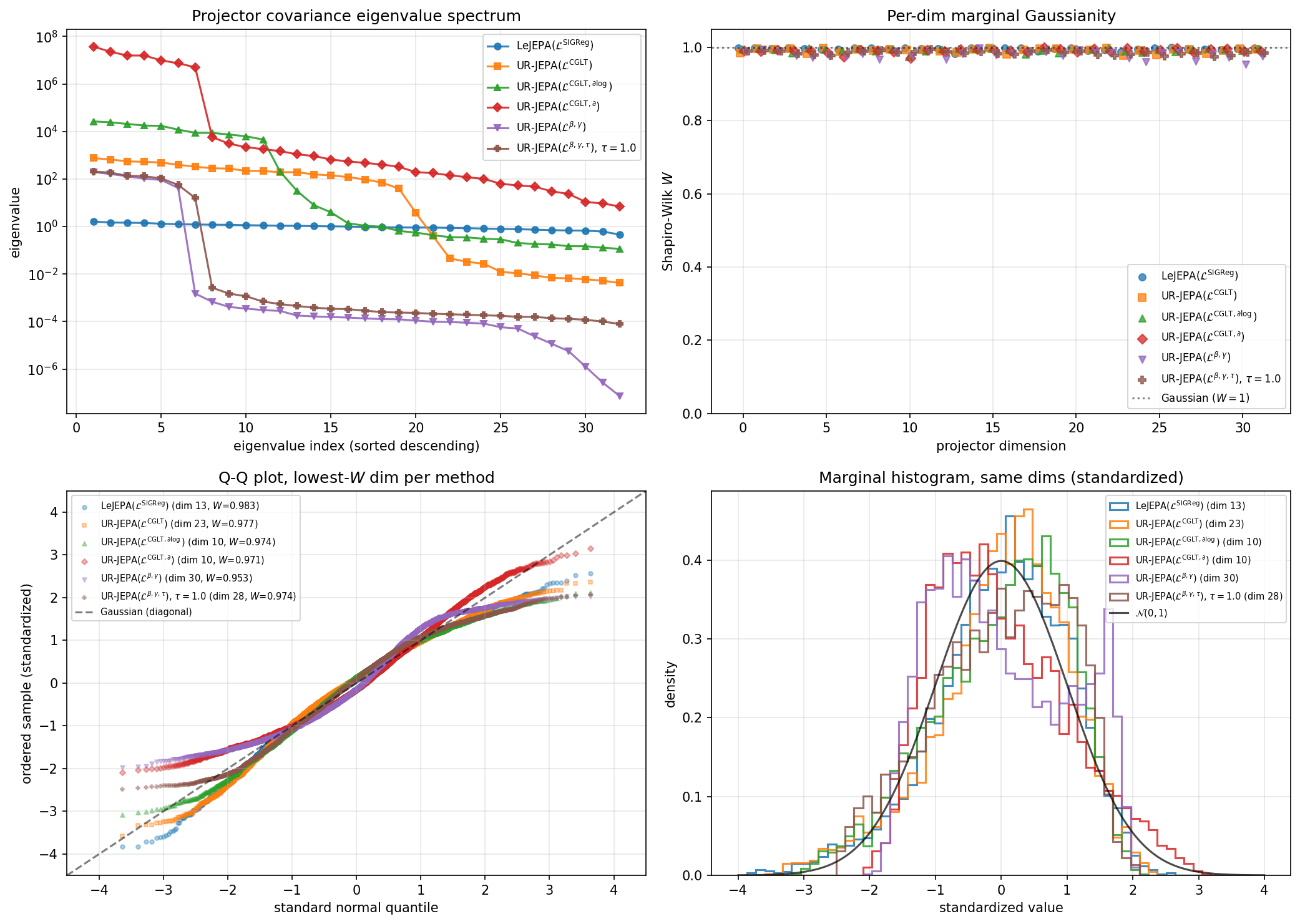}
\caption{Six-way projector-geometry comparison on Galaxy10~SDSS
at the seed-$0$ matched-recipe checkpoints of
Table~\ref{tab:galaxy10-matched}: LeJEPA($\Ll^{\text{SIGReg}}$) at
$D = 32$ together with the five UR--JEPA losses (integral
$\Ll^{\text{CGLT}}$, log-derivative $\Ll^{\text{CGLT},\partial\!\log}$,
raw-derivative $\Ll^{\text{CGLT},\partial}$, $\Ll^{\bet,\gamma}$,
and $\Ll^{\bet,\gamma,\tau}$ at $\tau = 1.0$). Panel layout matches
Figure~\ref{fig:viz-geometry-galaxy10}. LeJEPA is the sole
near-flat spectrum; all five UR--JEPA variants exhibit a cliff.
The cluster of methods into soft- and sharp-spectrum groups, and
the corresponding accuracy tiers of
Table~\ref{tab:galaxy10-matched}, is determined by the log
transform: $\Ll^{\text{CGLT},\partial\!\log}$ clusters with the
soft tier while $\Ll^{\text{CGLT},\partial}$ clusters with the
sharp tier.}
\label{fig:viz-geometry-galaxy10-6way}
\end{figure}

\begin{figure}[t]
\centering
\includegraphics[width=0.95\linewidth]{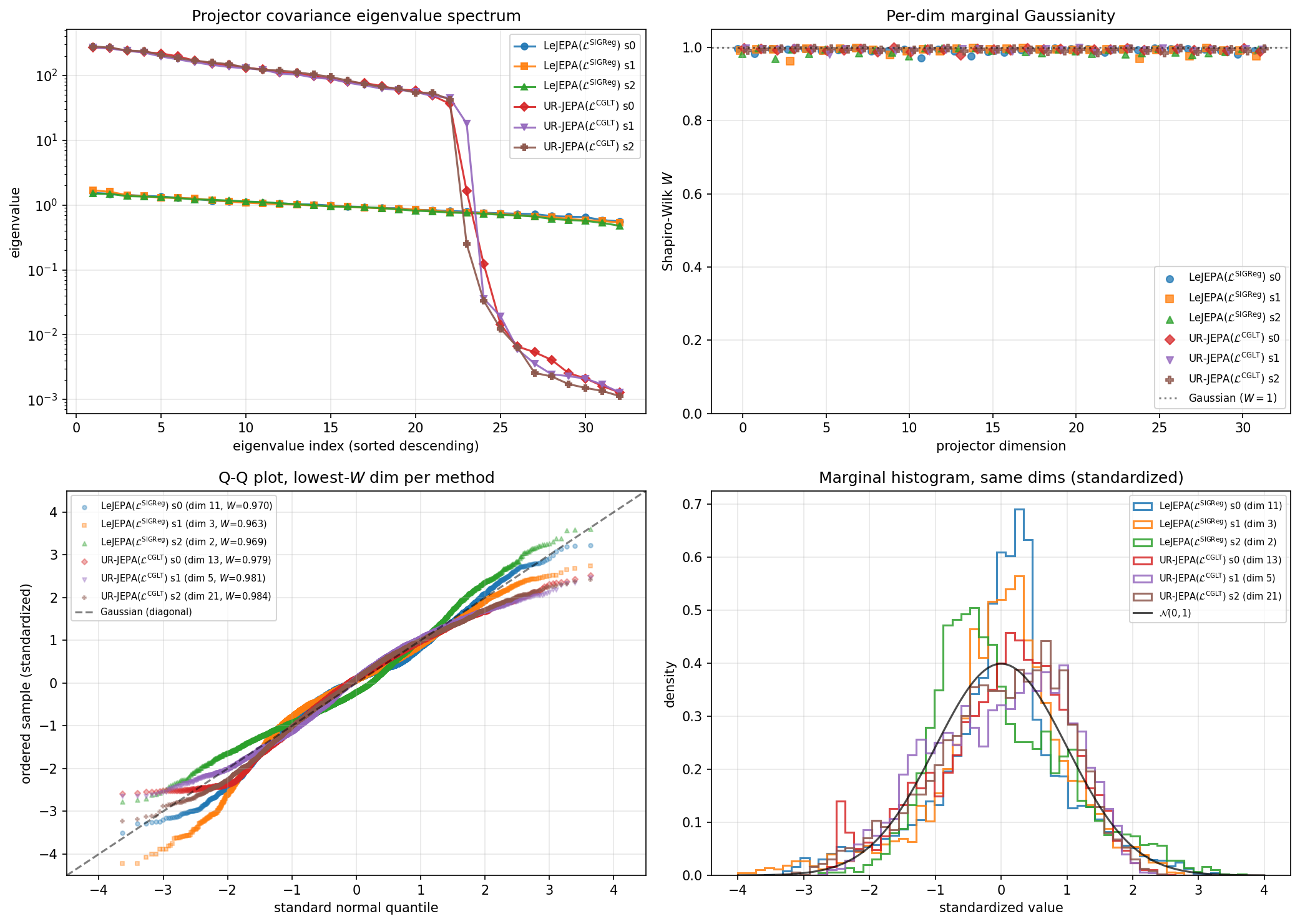}
\caption{Projector-geometry overlay for the EuroSAT
matched-recipe checkpoints: all $3$ seeds of each method
plotted together ($6$ traces). Panel layout matches
Figure~\ref{fig:viz-geometry-galaxy10-6way}. The three LeJEPA
traces (top of the spectrum panel) collapse onto a flat,
isotropic spectrum; the three UR--JEPA traces collapse onto a
shared anisotropic cliff at index $\sim\!22$ to $\sim\!24$.
The seed spread \emph{within} each method is negligible, so the
two methods form two tight, non-overlapping bundles --- a
multi-seed confirmation of the single-seed cliff signature
observed on the other three datasets.}
\label{fig:viz-geometry-eurosat-6way}
\end{figure}

\paragraph{Six-way extension on Galaxy10.}
Figure~\ref{fig:viz-geometry-galaxy10-6way} extends the two-method
comparison to all six matched-recipe variants of
Table~\ref{tab:galaxy10-matched} on Galaxy10~SDSS:
LeJEPA($\Ll^{\text{SIGReg}}$) at $D = 32$ together with the five
UR--JEPA losses (integral $\Ll^{\text{CGLT}}$, log-derivative
$\Ll^{\text{CGLT},\partial\!\log}$, raw-derivative
$\Ll^{\text{CGLT},\partial}$, $\Ll^{\bet,\gamma}$, and
$\Ll^{\bet,\gamma,\tau}$ at $\tau = 1.0$). Three observations are
visible. \textit{(i)} The covariance cliff is regularizer-family
deep, not method deep: LeJEPA yields the sole isotropic spectrum
(top-to-bottom ratio $3.6$); all five UR--JEPA variants exhibit a
cliff of at least $1.8 \times 10^{5}$. \textit{(ii)}
Theorem~\ref{thm:CGLT} equivalence extends to covariance space:
UR--JEPA($\Ll^{\text{CGLT}}$) and
UR--JEPA($\Ll^{\text{CGLT},\partial\!\log}$), formally equivalent
under Theorem~\ref{thm:CGLT}, have nearly identical spectrum shapes
(cliff position $\sim 20$ to $25$, top-to-bottom ratios
$1.8 \times 10^5$ vs $2.3 \times 10^5$). \textit{(iii)} The cosine
similarity of centered $\log$-eigenvalue profiles partitions the
six methods into two clusters whose membership matches the accuracy
tiers of Table~\ref{tab:galaxy10-matched}: a soft-spectrum cluster
(LeJEPA($\Ll^{\text{SIGReg}}$), $\Ll^{\text{CGLT}}$,
$\Ll^{\text{CGLT},\partial\!\log}$) with best accuracies in
$[0.8105, 0.8142]$, and a sharp-cliff cluster
($\Ll^{\text{CGLT},\partial}$, $\Ll^{\bet,\gamma}$,
$\Ll^{\bet,\gamma,\tau}$) with best accuracies in
$[0.7956, 0.7963]$. The cluster boundary is the log transform:
$\Ll^{\text{CGLT},\partial\!\log}$ sits in the soft-spectrum tier
while $\Ll^{\text{CGLT},\partial}$ sits in the sharp-cliff tier,
giving a covariance-side confirmation of the $+1.51$\,pp
accuracy gap attributable to the log transform
(\S\ref{sec:beta-rehab-galaxy10}). The $\beta$ family is anomalous
within the sharp-cliff cluster in that its leading eigenvalue is
small ($\sim 200$, compared with $\sim 7.5 \times 10^{2}$ to
$3.8 \times 10^{7}$ for the CGLT variants), so its anisotropy is a
tail-collapse story rather than a runaway leading direction; the
eig-threshold $\tau = 1.0$ lifts the tail floor by $\sim 10^{4}$
with no effect on the leading eigenvalue or on peak accuracy
($0.7956 \pm 0.0037$ vs $0.7958 \pm 0.0068$).

\paragraph{EuroSAT: $3$-seed confirmation of the cliff signature.}
The Inet10, Galaxy10, and Inet100 cells of
Table~\ref{tab:viz-eig-ratio} are single-seed. The EuroSAT
in-domain run of \S\ref{sec:eurosat} ($3$ seeds per method at
the matched recipe) provides the first multi-seed
confirmation that the cliff signature is reproducible across
seeds at fixed recipe. Figure~\ref{fig:viz-geometry-eurosat-6way}
overlays the projector-geometry diagnostics for all six
checkpoints ($3$ seeds $\times$ $2$ methods);
Table~\ref{tab:viz-eurosat} summarizes the per-seed numbers.
The two methods form tight, non-overlapping bundles: LeJEPA
holds an isotropic full-rank projector at all $3$ seeds
(effective rank $\sim\!29.5$, no cliff, all $D = 32$
dimensions alive); UR--JEPA produces a reproducible
anisotropic cliff at index $\sim\!22$ to $\sim\!24$ with
effective rank $\sim\!17.3$ at all $3$ seeds. Crucially, the
surviving dimensions stay near-Gaussian (min Shapiro-Wilk
$W \ge 0.979$ for UR--JEPA, with zero coordinates below
$W = 0.95$): the cliff is a clean rank reduction, not a
degenerate collapse.

\begin{table}[h]
\centering
\small
\setlength{\tabcolsep}{5pt}
\begin{tabular}{l c c c c}
\hline
method & effective rank & dims with norm-eig $> 10^{-4}$ & cliff index & min Shapiro $W$ \\
\hline
LeJEPA($\Ll^{\text{SIGReg}}$)
   & $29.7$\,/\,$29.3$\,/\,$29.4$
   & $32$\,/\,$32$\,/\,$32$
   & none (flat)
   & $0.970$\,/\,$0.963$\,/\,$0.969$ \\
UR--JEPA($\Ll^{\text{CGLT}}$)
   & $17.3$\,/\,$17.4$\,/\,$17.3$
   & $24$\,/\,$24$\,/\,$24$
   & $\sim\!22$\,/\,$23$\,/\,$22$
   & $0.979$\,/\,$0.981$\,/\,$0.984$ \\
\hline
\end{tabular}
\caption{EuroSAT projector-geometry $3$-seed diagnostics
(\texttt{ckpt-final.pt} from each seed, $5000$ validation
samples per checkpoint). Effective rank is the participation
ratio of the covariance eigenvalues; cliff index is the index
at which $\log_{10}(\mathrm{eig}_k / \mathrm{eig}_0)$ first
drops below $-2$; the final column reports the minimum
per-dimension Shapiro-Wilk $W$ across the $D = 32$ coordinates
(zero coordinates have $W < 0.95$ for either method).}
\label{tab:viz-eurosat}
\end{table}

\paragraph{Interpretation.}
At matched recipe and matched peak accuracy, the two regularizers
produce qualitatively different representations. LeJEPA targets
isotropic Gaussianity and achieves it; UR--JEPA targets local
uniform $n$-rectifiability and produces an effectively low-rank
projection (roughly $20$ to $25$ active dimensions out of $D = 32$)
while preserving the same per-dimension marginal shapes. The
spectral drop lies at index $\sim 20$ to $25$ rather than at the
target local tangent dimension $n = 7$, because
$\Ll^{\text{CGLT}}$ enforces \emph{local} $n$-rectifiability at
small scales $r$, not a global rank-$n$ constraint, so the global
covariance can have higher effective rank as long as small
neighborhoods look $n$-rectifiable. The position $\sim 20$ to $25$
is independently consistent with intrinsic-dimension estimates for
natural-image datasets reported by Pope~et~al.~\cite{PopeIntrinsic2021}
using the MLE and TwoNN estimators ($d \in [26, 43]$ for ImageNet,
$d \in [26, 35]$ for CIFAR-$10$), suggesting that the UR--JEPA
spectral drop tracks the dataset's intrinsic dimension rather than
the given target $n$. We emphasize that the diagnostic visualized
here is the \emph{global PCA rank} of the projector outputs: it
characterizes the ambient envelope of the projected cloud but does
not directly measure the local tangent dimension at small scales
that $\Ll^{\text{CGLT}}$ actually constrains.

\subsection{Summary of empirical findings}\label{sec:exp-summary}

\paragraph{The log transform is methodologically essential.}
UR--JEPA($\Ll^{\text{CGLT},\partial}$) (Thm.~\ref{thm:CGLT}\textnormal{(d)} via the
literal Eq.~(1.5) integrand, after the $t_{\max}$-rescaling of
\eqref{eq:Dtilde-rescaled}) attains $0.7963 \pm 0.0061$, a
$1.51$\,pp lower peak accuracy than
UR--JEPA($\Ll^{\text{CGLT},\partial\!\log}$) at the same
theoretical formulation. The combined seed-noise standard deviation
is $\sqrt{0.0037^2 + 0.0061^2} \approx 0.0071$, so a $1.51$\,pp
difference is more than $2 \sigma$ outside noise. The log-form
variants also exhibit lower peak-accuracy variance
($\pm 0.0017$, $\pm 0.0037$) than the raw form ($\pm 0.0061$).
This difference is direct empirical evidence that the log transform
of $\theta_t$ removes a parasitic local-density coupling and is
methodologically essential rather than a mere algebraic rewriting;
the ``The log-increment variant'' argument of \S\ref{sec:CGLTloss}
carries over from the dyadic-difference form to the scale-derivative
form.

The experiments support five claims:
\begin{enumerate}[leftmargin=2em,topsep=2pt,itemsep=2pt]
\item \textbf{Competitive performance with positive cross-$D$
scaling.} On Inet10 at the LeJEPA-exact recipe, the best
UR--JEPA($\Ll^{\text{CGLT}}$) configuration ($D = 32$, $n = 7$, $K = 5$) outperforms
matched-recipe LeJEPA($\Ll^{\text{SIGReg}}$) by $+0.83$~pp (paired-$t = +15.5$,
$p \ll 0.001$). Going from $D = 16$ to $D = 32$, LeJEPA($\Ll^{\text{SIGReg}}$) degrades
by $-0.72$~pp while UR--JEPA($\Ll^{\text{CGLT}}$) improves by $+0.19$~pp
(Table~\ref{tab:headline}).
\item \textbf{Lower seed variance.} The standard deviation of UR--JEPA($\Ll^{\text{CGLT}}$)
across seeds is $\sim$30\% smaller than LeJEPA($\Ll^{\text{SIGReg}}$)'s at the matched configuration.
\item \textbf{Wide $n$ plateau.} Probe accuracy varies by $< 1$~pp
across $n \in \{6, 7, 8, 9, 10\}$, falling off sharply only at the
collapse boundary $n \le 4$.
\item \textbf{Rehabilitation of the $\beta$-number variant.}
Although UR--JEPA($\Ll^{\bet}$) is known to collapse on Inet10 across
$5$ orders of magnitude of $s$, UR--JEPA($\Ll^{\bet,\gamma}$)
(pairing $\Ll_{\ur}^{\bet}$ with the intrinsic
$-\gamma \log \operatorname{tr}$ penalty of
\S\ref{sec:beta-anticollapse}) restores competitive performance on
Galaxy10~SDSS: $0.7970 \pm 0.0037$ over $3$ seeds at the A100
$\mathrm{bs} = 200$ recipe, $+0.76$\,pp above the matched-recipe
UR--JEPA($\Ll^{\text{CGLT}}$) at the same hardware tier
(\S\ref{sec:beta-rehab-galaxy10}, Table~\ref{tab:galaxy10-matched}).
The $\beta$ family is therefore alive on a non-trivial dataset
once paired with appropriate anti-collapse regularization, even
though the UR--JEPA($\Ll^{\bet}$) negative result is reproducible on
Inet10. UR--JEPA($\Ll^{\bet,\gamma,\tau}$) (adding the adaptive
eigenvalue-threshold rule on top of the log-trace penalty) does not
improve peak accuracy at $3$ seeds ($0.7958 \pm 0.0068$ at the
matched H100 recipe, statistically tied with
UR--JEPA($\Ll^{\bet,\gamma}$)); the eigenvalue threshold's value
lies in the conceptual properties documented in
\S\ref{sec:eig-thresh-discussion} rather than in peak accuracy.
\item \textbf{Theorem~\ref{thm:CGLT} equivalence holds in training,
log form only.} The matched-recipe four-variant Galaxy10 study
(Table~\ref{tab:galaxy10-matched}) confirms that
UR--JEPA($\Ll^{\text{CGLT}}$) (dyadic-difference \textnormal{(b)})
and UR--JEPA($\Ll^{\text{CGLT},\partial\!\log}$) (log
scale-derivative \textnormal{(d)}) are training-equivalent on peak
accuracy: $0.8142 \pm 0.0017$ vs.\ $0.8114 \pm 0.0037$, a
$0.28$\,pp gap below the seed-noise floor of $\sqrt{0.0017^2 +
0.0037^2} \approx 0.0041$. UR--JEPA($\Ll^{\text{CGLT},\partial}$)
(the literal-raw form of (d)) underperforms by $1.51$\,pp
($0.7963 \pm 0.0061$), over
$2 \sigma$ outside noise: direct empirical evidence that the log
transform of $\theta_t$ in \S\ref{sec:CGLTloss} removes a parasitic
density-level coupling and is methodologically essential.
\end{enumerate}
Taken together, claims (1)--(3) establish UR--JEPA($\Ll^{\text{CGLT}}$) as the validated
default at the matched LeJEPA recipe; claim (4) positions
UR--JEPA($\Ll^{\bet,\gamma}$) as a viable practical alternative on a
non-trivial dataset, particularly well-suited to the high
ambient-dimensional regime where density-based estimators become
noisy; and claim (5) confirms that UR--JEPA($\Ll^{\text{CGLT}}$) and
UR--JEPA($\Ll^{\text{CGLT},\partial\!\log}$) are training-equivalent in practice,
validating UR--JEPA($\Ll^{\text{CGLT},\partial\!\log}$) as a deployment-ready
alternative with $4.8 \times$ tighter late-training stability than
UR--JEPA($\Ll^{\text{CGLT}}$). On Galaxy10 specifically, UR--JEPA($\Ll^{\text{CGLT}}$),
UR--JEPA($\Ll^{\text{CGLT},\partial\!\log}$), and the matched-recipe LeJEPA($\Ll^{\text{SIGReg}}$) baseline
are within $\sim 0.37$\,pp of each other on peak accuracy: most of
the $+6.10$\,pp gap to LeJEPA's published Galaxy10 result is
recipe-attributable, not regularizer-attributable, and the
Galaxy10 finding is more accurately stated as ``in-domain SSL
outperforms foundation-model transfer by $\ge 18$\,pp'' rather than
as a loss-family advantage.

\section{Extending LeJEPA to uniform rectifiability: discussion}\label{sec:discussion}

\subsection{Structural correspondence with LeJEPA}
LeJEPA's framework decomposes cleanly as
\[
\underbrace{\Ll_{\pred}}_{\text{invariance}}
\;+\;
\lambda\underbrace{\Ll_{\sig}}_{\text{shape of marginal law}}.
\]
UR--JEPA preserves this decomposition and only substitutes the shape
term:
\[
\underbrace{\Ll_{\pred}}_{\text{invariance}}
\;+\;
\underbrace{\lambda_1 \Ll_{\ur} + \lambda_2 \Ll_{\ad}}_{\text{shape of marginal law}}.
\]
The substitution targets a different geometric hypothesis. We ask for
quantitatively $n$-dimensional support at all scales and locations,
instead of a full-dimensional Gaussian. Note that the Carleson-measure
formulation of UR is the natural analogue of LeJEPA's sketched
Gaussianity test. Indeed, both are sums of scale-local defect
statistics that vanish exactly on the target class.

\subsection{What does UR--JEPA recover that LeJEPA cannot?}
\begin{enumerate}[leftmargin=*]
\item \emph{Local tangent dimension as a tunable parameter.} The
target local tangent dimension $n$ at small scales is given as
input; LeJEPA has no analogous parameter. What is empirically
demonstrated in \S\ref{sec:viz-geometry} is an effectively
low-rank projector envelope at the global PCA level, consistent
with the data's intrinsic-dimension estimates; a direct
verification that the local tangent dimension at small scales is
in fact $\sim n$ requires a per-anchor local-effective-rank
measurement and is flagged as a follow-up
(\S\ref{sec:future}). The cost of the extra hyperparameter is
small in practice: probe accuracy varies by under $1$\,pp across
$n \in \{6, 7, 8, 9, 10\}$ at $D = 16$ (\S\ref{sec:ablation-n}).

\item \emph{Scale-free shape constraints.} Theorem~\ref{thm:CGLT}
enforces a condition that is Carleson-type in $(x, r)$, that is, it
constrains the geometry at every location and every scale
simultaneously, rather than enforcing a single global distributional
target.

\item \emph{Compatibility with structured supports.} Uniformly
$n$-rectifiable measures include measures supported on smooth
$n$-manifolds, finite unions of such, Lipschitz graphs, and more.
LeJEPA implicitly forbids these because they have zero volume in
$\R^D$.
\end{enumerate}

The empirical consequences (an effectively low-rank projector
spectrum at matched recipe, marginal Gaussianity preserved as a
Diaconis-Freedman consequence, lower seed variance, and mild
sample-efficiency gains) are documented in
\S\ref{sec:viz-geometry} and synthesized in
\S\ref{sec:exp-summary}.

\subsection{What does UR--JEPA sacrifice?}
\begin{enumerate}[leftmargin=*]
\item \emph{Closed-form optimum.} LeJEPA identifies
$\mathcal{N}(0, I_D)$ as Bayes-optimal for downstream prediction under
its assumptions. We do not have the analogous result for UR--JEPA.
Recall that UR is a class of target measures, not a single
distribution, and the choice among them is left to $\Ll_{\pred}$.
\item \emph{One extra hyperparameter} ($n$), plus a second coefficient
for $\Ll_{\ad}$ in the CGLT variant.
\item \emph{Higher per-step cost} for the CGLT loss than for SIGReg
(pairwise distances versus sketched projections). The $\beta$ variant
is competitive in flops for reasonable anchor counts.
\end{enumerate}

\section{Summary}\label{sec:summary}

Replacing SIGReg's Gaussian target by a uniform-rectifiability
target gives a one-to-one substitution inside the LeJEPA recipe and
yields two families of concrete, differentiable losses:
\[
\boxed{\;
\Ll^{\text{CGLT},\,\star} = \Ll_{\pred} + \lambda_1 \Ll_{\ur}^{\text{CGLT},\,\star} + \lambda_2 \Ll_{\ad}
\qquad\text{or}\qquad
\Ll^{\bet,\,\star} = \Ll_{\pred} + \lambda\, \Ll_{\ur}^{\bet,\,\star}
\;}
\]
where $\Ll_{\ur}^{\text{CGLT},\,\star}$ denotes any of three concrete
Carleson square-functions on the Gaussian-smoothed density:
\eqref{eq:L-CGLT} (dyadic difference of $\log \theta_r$, the default),
\eqref{eq:L-CGLT-deriv} (log-derivative $t\, \partial_t \log \theta_t$),
or \eqref{eq:L-CGLT-deriv-raw} (the literal Eq.~(1.5) of
\cite{CGLT2014}). $\Ll_{\ur}^{\bet,\,\star}$ denotes the
$\beta$-number Carleson sum \eqref{eq:L-beta} augmented by one of the
two anti-collapse mechanisms of \S\ref{sec:beta-anticollapse}
(log-trace penalty or adaptive eigenvalue-threshold tangent
selection). Both families are consistent with the corresponding
characterization of uniform $n$-rectifiability in
Theorem~\ref{thm:CGLT} and the
$\beta$-number counterpart \eqref{eq:betaCarleson} under the
empirical measure, and both admit efficient SGD implementations.
Empirically (\S\ref{sec:experiments}), UR--JEPA($\Ll^{\text{CGLT}}$)
is the validated default; it exceeds matched-recipe
LeJEPA($\Ll^{\text{SIGReg}}$) on Inet10 by $+0.83$\,pp at $3$
seeds (paired-$t = +15.5$, $p \ll 0.001$) with $\sim\!30\%$ lower
seed standard deviation, and on Galaxy10~SDSS, Inet100, and
EuroSAT the two methods lie in the same peak-accuracy band at
convergence. UR--JEPA($\Ll^{\bet}$) collapses on Inet10 across
$5$ orders of magnitude of $s$, but UR--JEPA($\Ll^{\bet,\gamma}$)
makes the $\beta$ family viable on Galaxy10~SDSS. On Galaxy10~SDSS, UR--JEPA($\Ll^{\text{CGLT}}$)
and UR--JEPA($\Ll^{\text{CGLT},\partial\!\log}$) are statistically tied on peak
accuracy at $3$ seeds (Theorem~\ref{thm:CGLT} equivalence holding in
training), with UR--JEPA($\Ll^{\text{CGLT},\partial\!\log}$) offering $4.8 \times$
tighter late-training stability. UR--JEPA($\Ll^{\text{CGLT},\partial}$) underperforms
UR--JEPA($\Ll^{\text{CGLT},\partial\!\log}$) by $1.51$~pp, evidence that the log
transform is methodologically essential rather than a mere
algebraic rewriting.

\section{Future work}\label{sec:future}

Several directions follow naturally from the present study.

\paragraph{Remote sensing, multimodal, and multisource data.}
Our experiments span Inet10 ($\sim 13\mathrm{K}$ images),
Galaxy10~SDSS ($\sim 21\mathrm{K}$), and single-seed Stage-$2$
ImageNet-$100$ ($\sim 130\mathrm{K}$). The natural next domain for
UR--JEPA is remote sensing: satellite and aerial imagery are
structurally biased toward low-dimensional regularities (land-cover
textures, sensor noise, geographic priors) where the
rectifiable-support hypothesis underpinning UR--JEPA is plausibly
sharper than on ImageNet-style scenes. Standard benchmarks
(EuroSAT, BigEarthNet, fMoW, SpaceNet) provide a graded difficulty
ladder. Multimodal extensions are a logical next step: optical, SAR,
multispectral, hyperspectral, and elevation modalities each project
to a different rectifiable structure, so a joint embedding obtained
by applying $\Ll^{\text{CGLT}}$ within each modality and an
invariance term across modalities is a principled formulation. The
multisource setting (sensor and platform heterogeneity, varying
acquisition geometry, mixed labeling regimes) is a further test of
whether a single geometric loss family can absorb the distributional
shifts that complicate distribution-matching objectives such as
SIGReg.

\paragraph{Theoretical analysis of UR--JEPA.}
The present work motivates UR--JEPA from the CGLT and Jones-$\beta$
characterizations of uniform $n$-rectifiability, but does not
provide a finite-sample convergence guarantee. Open questions
include: (a) under what conditions does minimizing
$\Ll^{\text{CGLT}}$ provably drive the projected distribution
toward an $n$-AD-regular, uniformly $n$-rectifiable target?
(b) how does the sample complexity scale with the ambient projector
dimension $D$, the target intrinsic dimension $n$, and the number
of dyadic scales $K$? (c) can the CGLT square function be
connected, via a gradient-flow or neural-tangent-kernel argument,
to the projected distribution's local-PCA spectrum? A
second-moment analysis at the projector output is a natural
starting point.

\paragraph{Downstream tasks beyond classification.}
The downstream evaluation in \S\ref{sec:inet100} restricts
attention to linear-probe classification on object-centric
datasets. Dense prediction tasks (semantic segmentation, instance
detection) place direct demands on the spatial-local structure of
the learned features, which is precisely what UR--JEPA targets
through its localized geometric statistics. Whether the
matched-recipe convergence parity observed on classification
persists under dense prediction, or whether one regularizer family
pulls ahead, is open.

\paragraph{Adaptive scale ladder and projector design.}
The dyadic ladder $K = 5$ used throughout is a hand-picked default;
the $K$-sweep at the best $n$ on each dataset is deferred. A
data-driven choice of the dyadic step (e.g., adapting to the
per-minibatch pairwise-distance distribution) and a systematic
study of the projector hidden-norm and width are natural
follow-ups.

\paragraph{Alternative schedules for the intrinsic dimension $n$.}
Two schedules beyond fixed $n$ are worth evaluating but were not
tested here: \emph{curriculum} $n$ (annealing from a large value
down to the target to avoid early trapping in degenerate flat
configurations) and \emph{learned soft} $n$ (replacing the hard
top-$n$ partition by a soft attention over eigenvalues with a
learned temperature). The latter introduces an extra hyperparameter
and somewhat undermines the dimensional-homogeneity argument, so
its empirical case is less clear than for curriculum $n$.

\section*{Code availability}

Code to reproduce the experiments in this paper will be made
available at \url{https://github.com/SPATIOLYX/UR-JEPA}.

\section*{Acknowledgements}

The author gratefully acknowledges the National Science Foundation
ACCESS program\footnote{\url{https://access-ci.org/}} for awarded
compute allocations, and the Anvil cluster operated by the Rosen
Center for Advanced Computing at Purdue
University\footnote{\url{https://www.rcac.purdue.edu/anvil}} for
providing the GPU resources on which the experiments in this paper
were carried out. The author also thanks Professor Gilad Lerman for
useful feedback on the $\bet$-number formulation.

\end{document}